\documentclass{article} % For LaTeX2e
\usepackage{iclr2026_conference,times}
\usepackage[final]{changes}
\usepackage{amsmath,amsfonts,bm}

\def\eqref#1{equation~\ref{#1}}
\def\1{\bm{1}}

\DeclareMathAlphabet{\mathsfit}{\encodingdefault}{\sfdefault}{m}{sl}
\SetMathAlphabet{\mathsfit}{bold}{\encodingdefault}{\sfdefault}{bx}{n}

\usepackage{hyperref}
\usepackage{url}
\usepackage{booktabs}       % professional-quality tables
\usepackage{amsfonts}       % blackboard math symbols
\usepackage{nicefrac}       % compact symbols for 1/2, etc.
\usepackage{microtype}      % microtypography
\usepackage{graphicx}
\usepackage{subcaption}
\usepackage{amsmath,epstopdf,amssymb,color,url,multirow,multicol,verbatim,threeparttable,diagbox,makecell,booktabs,color,colortbl}

\usepackage{algorithm}
\usepackage{algpseudocode}

\usepackage[dvipsnames]{xcolor}
\definecolor{light-gray}{rgb}{0.9, 0.95, 1}

\usepackage{adjustbox}
\usepackage{amsmath}
\usepackage{wrapfig}
\usepackage{mathtools}

\title{All Patches Matter, More Patches Better:  \\ Enhance AI-Generated Image Detection via Panoptic Patch Learning}

\author{
  \vspace{-25pt}\\
  \textbf{Zheng Yang$^{1}$\thanks{: Equal Contribution, $\dag$: Corresponding Author},\quad Ruoxin Chen$^{2\star \dag}$,\quad Zhiyuan Yan$^{3}$,\quad Keyue Zhang$^2$,\quad Xinghe Fu$^1$}, \\
  \textbf{Shuang Wu$^2$,\quad Xiujun Shu$^4$,\quad  Taiping Yao$^2$,\quad Shouhong Ding$^2$,\quad Zequn Qin$^{5\dag}$,\quad Xi Li$^{1\dag}$}\vspace{3pt} \\
  $^1$College of Computer Science and Technology, Zhejiang University \\ $^2$Youtu Lab, Tencent ~~\quad
  $^3$Peking University ~~\quad $^4$Wechat Pay, Tencent \\ $^5$School of Software Technology, Zhejiang University\vspace{3pt} \\
  \vspace{-25pt} \\
}

\iclrfinalcopy % Uncomment for camera-ready version, but NOT for submission.
\begin{document}

\maketitle

\begin{abstract}
The rapid proliferation of AI-generated images (AIGIs) highlights the pressing demand for generalizable detection methods. In this paper, we establish two key principles for AIGI detection task through systematic analysis:
\textbf{(1) All Patches Matter},  
since the uniform generation process ensures that each patch inherently contains synthetic artifacts, making every patch a valuable detection source; and
\textbf{(2) More Patches Better}, as leveraging distributed artifacts across more patches improves robustness by reducing over-reliance on specific regions.
However, counterfactual analysis uncovers a critical weakness: naively trained detectors display \textbf{Few-Patch Bias}, relying disproportionately on \emph{minority patches}.
We identify this bias to \textbf{Lazy Learner} effect, where detectors to limited patch artifacts while neglecting distributed cues.
To address this, we propose \textbf{Panoptic Patch Learning} framework, which integrates: 
(1) \emph{Randomized Patch Reconstruction}, injecting synthetic cues into randomly selected patches to diversify artifact recognition; 
(2) \emph{Patch-wise Contrastive Learning}, enforcing consistent discriminative capability across patches to ensure their uniform utilization.
Extensive experiments demonstrate that PPL enhances generalization and robustness across datasets.
\end{abstract}

\section{Introduction}
\label{sec:intro}

The rapid evolution of generative AI models has precipitated an exponential growth of AI-generated images (AIGIs) in digital ecosystems~\citep{goodfellow2014generative,karras2018progressive,karras2019style,ho2020denoising,rombach2022high,zhang2023controlvideo,pmlr-v139-ramesh21a,yan2025gpt,yan2024df40,yan2025unified,zhang2025easycontrol,zhang2024ssr,song2025makeanything,song2025layertracer}. This proliferation raises concerns regarding information security and content authenticity, highlighting the need for AIGI detection to distinguish synthetic images from authentic ones. Unlike conventional classification tasks, AIGI detection operates as a ``\emph{cat-and-mouse game}'', presenting unique challenges due to: (1) continuous emergence of new generative architectures, and (2) frequent updates to existing generative models. Consequently, exhaustive training on all synthetic data becomes impractical~\citep{ojha2023towards}, thus necessitating detectors with strong generalizability.

Despite these challenges, AIGIs exhibit a distinctive property absent in traditional classification: Universal Artifact Distribution. In the context of AIGIs, discriminative features are not confined to object-centric regions; instead, \emph{synthetic images contain artifacts uniformly across all patches, a consequence of the consistent generation process of modern generative models}.\footnote{This work adheres to the mainstream AIGI detection setting~\citep{chen2024drct,ojha2023towards,tan2024c2p,liu2024forgery,he2024rigid,zhu2024genimage,tan2024rethinking,lin2025seeing} where the entire image is generated by AI models.} 
% Unlike conventional classification tasks, where discriminative features are concentrated in regions containing the label objects, synthetic images exhibit artifacts uniformly distributed across all patches due to the generative models' uniform production process. 
This indicates that every patch, \added{defined as partitioned local sub-blocks}, contains synthetic traces, forming our first principle for AIGI detection: All Patches Matter. This principle is supported by two lines of evidence: (1) visual analytics~\citep{tan2024rethinking,cozzolino2024zero,chen2025dual} confirm pixel-level discriminative patterns, revealing artifacts at patch granularity; and (2) recent patch-wise detectors~\citep{chen2024single,zhong2024patchcraft} demonstrate comparable performance to full-image approaches, validating the discriminative capability of individual patches. Although artifacts vary across patches, detectors that capture diverse artifacts across distributed regions reduce over-reliance on specific patches. Capturing these distributed artifacts enhances cross-generator generalizability by mitigating detectors’ blind spots. This leads to our second principle: More Patches Better.

\begin{figure}[tp!]
\centering
\subfloat[Visualization of attention maps.]{\includegraphics[height=6cm,width=6cm]{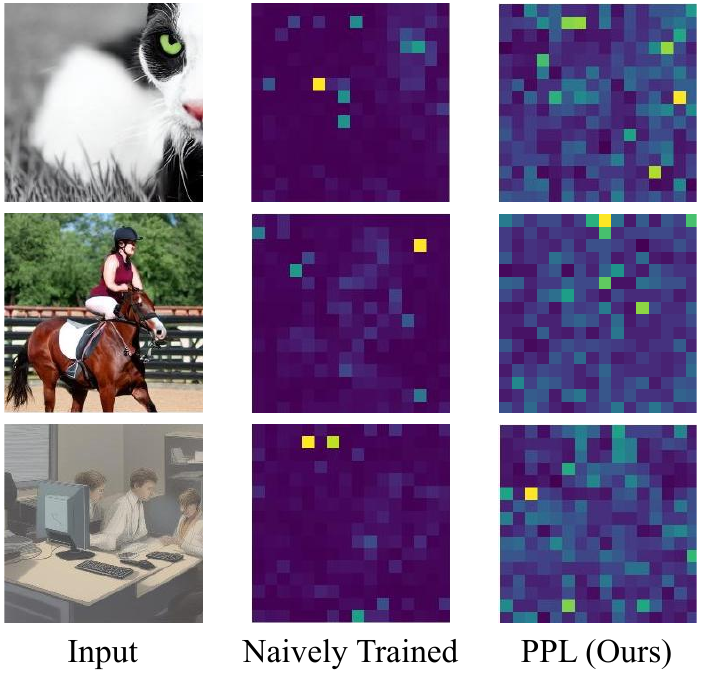}}
\hspace{0.5cm}
\subfloat[Comparison on different datasets.]{\includegraphics[height=6.5cm,width=6.5cm]{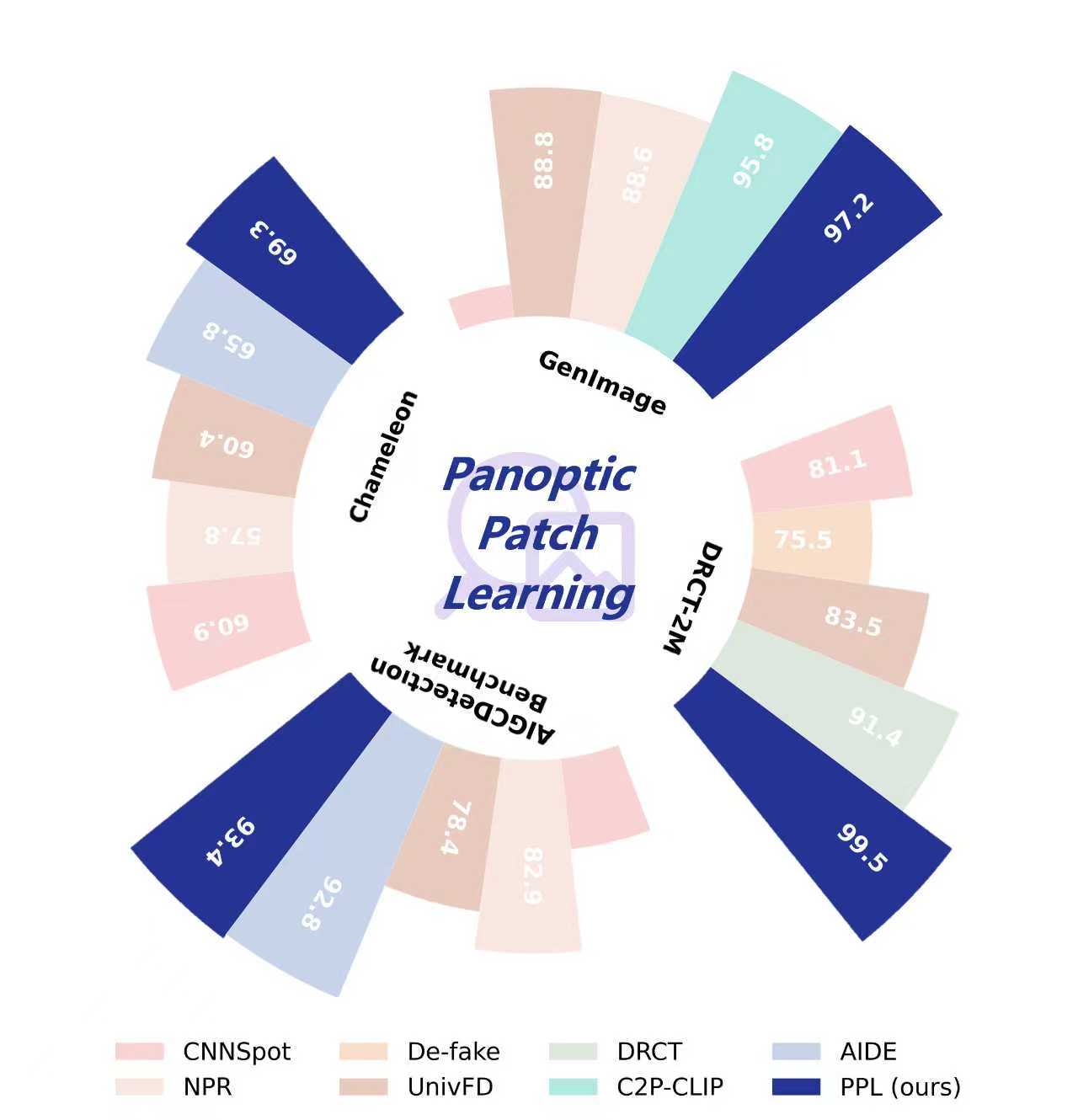}}
\caption{\textbf{(a)} PPL produces a more uniform attention distribution across patches, indicating its effectiveness in capturing artifacts comprehensively. \textbf{(b)} PPL outperforms peer methods on GenImage, DRCT-2M, AIGCDetectionBenchmark, and the in-the-wild Chameleon. More details are in Section~\ref{sec:exper}.}
\label{fig:intro}
\end{figure}

However, counterfactual analysis of existing AIGI detectors~\citep{ojha2023towards,liu2024forgery,tan2024c2p,he2024rigid,chen2024drct} reveals an unfavorable tendency—Few-Patch Bias—supported by two empirical observations and a quantitative analysis. Empirically, we observe: (1) detectors' attention maps disproportionately focus on very limited patches; (2) detectors exhibit severe patch-specific fragility, where masking a single patch reduces accuracy by 18.7\% $\pm$ 4.1\% on average. Quantitatively, using causal inference tool \added{CDE}~\citep{vanderweele2013three} to quantify each patch's impact—measured as the classification logit difference with and without that patch—we find that naively trained detectors produce skewed distributions: a few patches show high \added{CDE} values, while most patches exhibit significantly lower contributions. This suggests that most patches remain underutilized, despite also containing generative artifacts. Moreover, \emph{detection methods with more uniform \added{CDE} distributions exhibit stronger generalizability}; for instance, DRCT, with more high-\added{CDE} patches, performs substantially better than UnivFD. We attribute such \added{\emph{Few-Patch Bias}} to the propensity of detectors as \emph{Lazy Learner}~\citep{hermann2024foundation,zhang2021if,wang2022frequency,zhao2024comi,ghosh2023tackling,tang2023large,sun2024exploring,yuan2024llms,yan2024effort}. 
Specifically, AIGI detectors follow a curriculum-like learning pattern: \emph{once easy-to-learn artifacts in certain patches minimize loss, the presence of these patches discourages exploration of broader regions.}

% a fundamental inconsistency with ‘All Patches Matter’ principle. Two key findings emerge:  Through causal analysis using Controlled Direct Effect (\added{CDE})~\citep{vanderweele2013three}, we quantify patch impacts by measuring logit differences before and after masking, higher \added{CDE} scores mean higher impact. Notably, The \added{CDE} distribution of image patches reveals a heavy-tailed pattern where 78.3\% of patches show negligible impact (\added{CDE} < 0.1) while only 4.2\% account for 63\% of total impact. Empirical evidence indicates that this minority-driven discrimination paradigm overly relies on a limited number of high-\added{CDE} patches for real-synthetic discrimination while underutilizing synthetic artifacts in non-salient regions. This ‘few patches matter’ paradigm contradicts our ‘all patches matter’ principle. Importantly methods demonstrating higher patch utilization (more high-\added{CDE} patches) achieve superior performance validating that broader artifact exploitation enhances discrimination capability.

To address this challenge, we propose the principle: ``All Patches Matter, More Patches Better", which prevents detectors from shortcutting to a few regions and instead encourages robust feature learning across the entire image. To operationalize this principle, we introduce the \textbf{Panoptic Patch Learning} (PPL) framework, which consists of two components: 
(1) \emph{Randomized Patch Reconstruction}, which manually injects synthetic artifacts into randomly selected patches of real images via diffusion reconstruction, forcing the model to discriminate based on these chosen regions and discouraging over-reliance on specific patches; and 
(2) \emph{Patch-wise Contrastive Learning}, which aligns the representations of real and synthetic patches, thereby enforcing consistent discriminative capability across all regions of the image.
Fig.~\ref{fig:intro} illustrates the effectiveness of PPL. Our main contributions are threefold:
\begin{enumerate}
    \item We formally propose the principle ``All Patches Matter, More Patches Better", showing that exploiting distributed artifacts enhances AIGI detection.
    \item We provide a detailed patch-wise analysis using \added{CDE}, revealing that Few-Patch Bias is pervasive in existing detectors.
    \item Building on this principle, we design Panoptic Patch Learning and validate its effectiveness through extensive experiments.
\end{enumerate}

\section{Related Work}
\label{sec:rela}

Existing AIGI detection methods can be broadly categorized into two types: \emph{local} and \emph{global} detection~\citep{tan2024c2p}. We summarize both lines of research below.

\paragraph{Local AIGI detection methods.}  
Local approaches exploit localized information to distinguish AI-generated images from real ones, assuming that low-level feature differences exist between the two. These methods can be divided into \emph{patch-wise} and \emph{pixel-wise} detectors.  

\emph{Patch-wise methods} include:  
SSP~\citep{chen2024single} achieves notable performance using only a single patch.
Patchcraft~\citep{zhong2024patchcraft} separates processing of the simplest and most complex patches by entropy-based selection.
~\citep{zheng2024break} employ a patch-based CNN leveraging all patches to avoid selective sampling and aggregate patch features.  
TextureCrop~\citep{konstantinidou2025texturecrop} partitions an image via sliding windows and selects high-frequency texture-rich regions.  
Despite these advances, patch-wise detectors often over-rely on a limited subset of patches, leading to information under-utilization. \emph{Pixel-wise methods} include:  
NPR~\citep{tan2024rethinking} detects AIGIs by analyzing differences in neighboring pixel relationships. 
FreqNet~\citep{tan2024frequency} and SAFE~\citep{li2024improving} exploit high-frequency signals to capture localized patterns.  
% ZED~\citep{cozzolino2024zero}, which measures the coding cost of local regions via an entropy-based encoder and identifies forgeries by detecting gaps in coding costs.  
However, pixel-wise methods are sensitive to small perturbations in pixel relationships, limiting their robustness.  

\paragraph{Global AIGI detection methods.}  
Global approaches leverage holistic image characteristics to distinguish AIGIs from real images, aiming to capture inconsistencies that may not be observable at the local level.  
CNNSpot~\citep{wang2020cnn} applies a CNN directly for detection, achieving strong in-distribution performance but suffering from poor cross-generator generalization.  
UnivFD~\citep{ojha2023towards} improves robustness by adopting a CLIP visual encoder as a feature extractor.  
FatFormer~\citep{liu2024forgery} further adapts CLIP by introducing a frequency adapter.  
C2P-CLIP~\citep{tan2024c2p} fine-tunes CLIP with carefully designed image–text pairs to embed the notions of “real” and “fake.”  
DRCT~\citep{chen2024drct} strengthens UnivFD with a contrastive loss on hard cases.  
Nevertheless, global methods often overlook fine-grained forensic artifacts, which constrains their effectiveness.

\section{Motivation}
\label{sec:motiv}
\subsection{All Patches Matter, More Patches Better}
\begin{figure}[t!]
    \centering
    \begin{adjustbox}{width=0.9\linewidth}
    \includegraphics{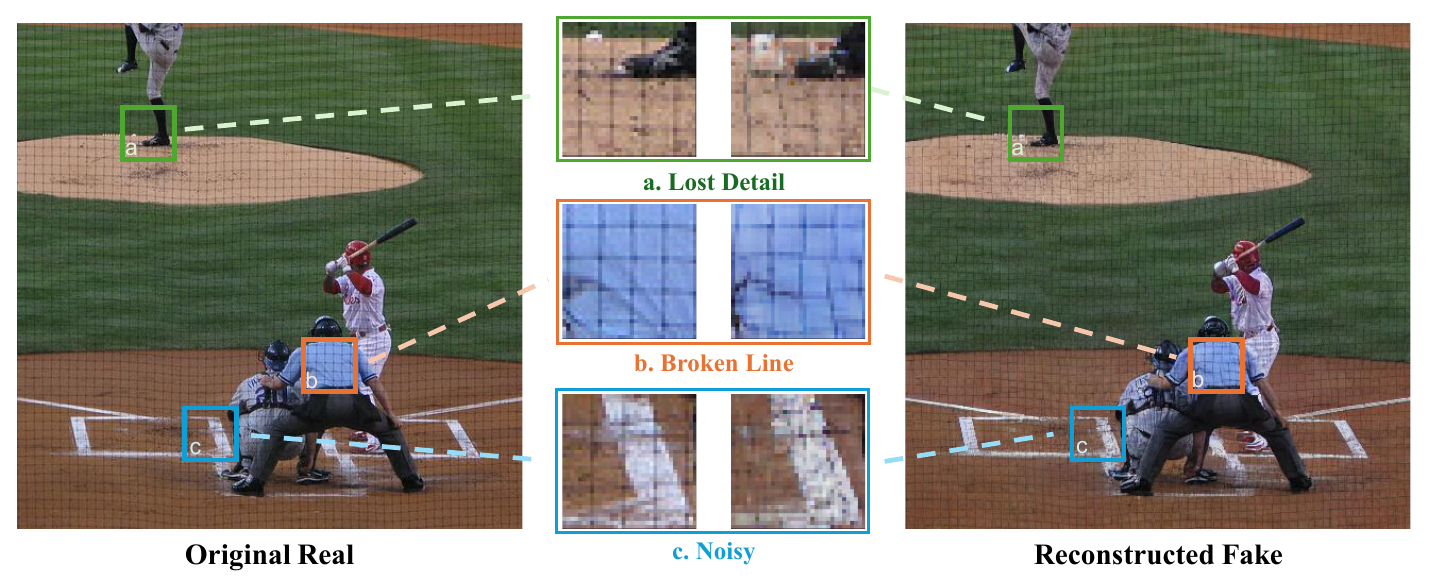}
    \end{adjustbox}
    \caption{Visual evidence of patch-wise artifacts. We observe diverse traces—such as broken lines, unnatural noise, and boundary detail loss—showing that multiple regions of synthetic images contain cues. This observation underscores the importance of \emph{leveraging more patches to enhance recognition of diverse artifacts}. Images are sourced from MSCOCO~\citep{lin2014microsoft}.}
    \label{fig:pattern}
    \vspace{-5pt}
\end{figure}

The principle of \emph{All Patches Matter} is supported by three key findings.  
\begin{enumerate}
    \item \textbf{\added{Principle:}} Because every patch of a synthetic image is itself generated, each inherently contains artifacts. Localized detection methods~\citep{chen2024single, zhong2024patchcraft} demonstrate that cues within small regions can effectively discriminate real from synthetic content, underscoring that every patch carries discriminative signals.  
    \item \textbf{Visualization:} Fig.~\ref{fig:pattern} illustrates distinct artifact patterns across patches, showing that each synthetic patch exhibits identifiable features distinguishing it from real patches. Moreover, the variability of these cues across patches highlights the considerable diversity of artifacts present in synthetic images.  
    \item \textbf{Experiments:} We further validated this principle by evaluating detectors on single randomly selected patches. By replicating one patch across the image to isolate its features, detectors still achieved 90\% accuracy on the SDv1.4 subset of GenImage. This confirms that even a single patch contains sufficient information for reliable discrimination.  
\end{enumerate}
Together, these findings demonstrate that artifacts in synthetic images are both pervasive and diverse. Detectors can exploit these patch-level cues, motivating the principle of \emph{More Patches Better}: leveraging more patches enhances robustness and generalization by capturing complementary artifact patterns. However, our observations reveal that existing detectors \added{do not} align with this principle.

\subsection{Few-Patch Bias}
\begin{figure}[t!]
\centering
\subfloat[Visualization of Attention Maps.]{\includegraphics[height=5cm,width=7.5cm]{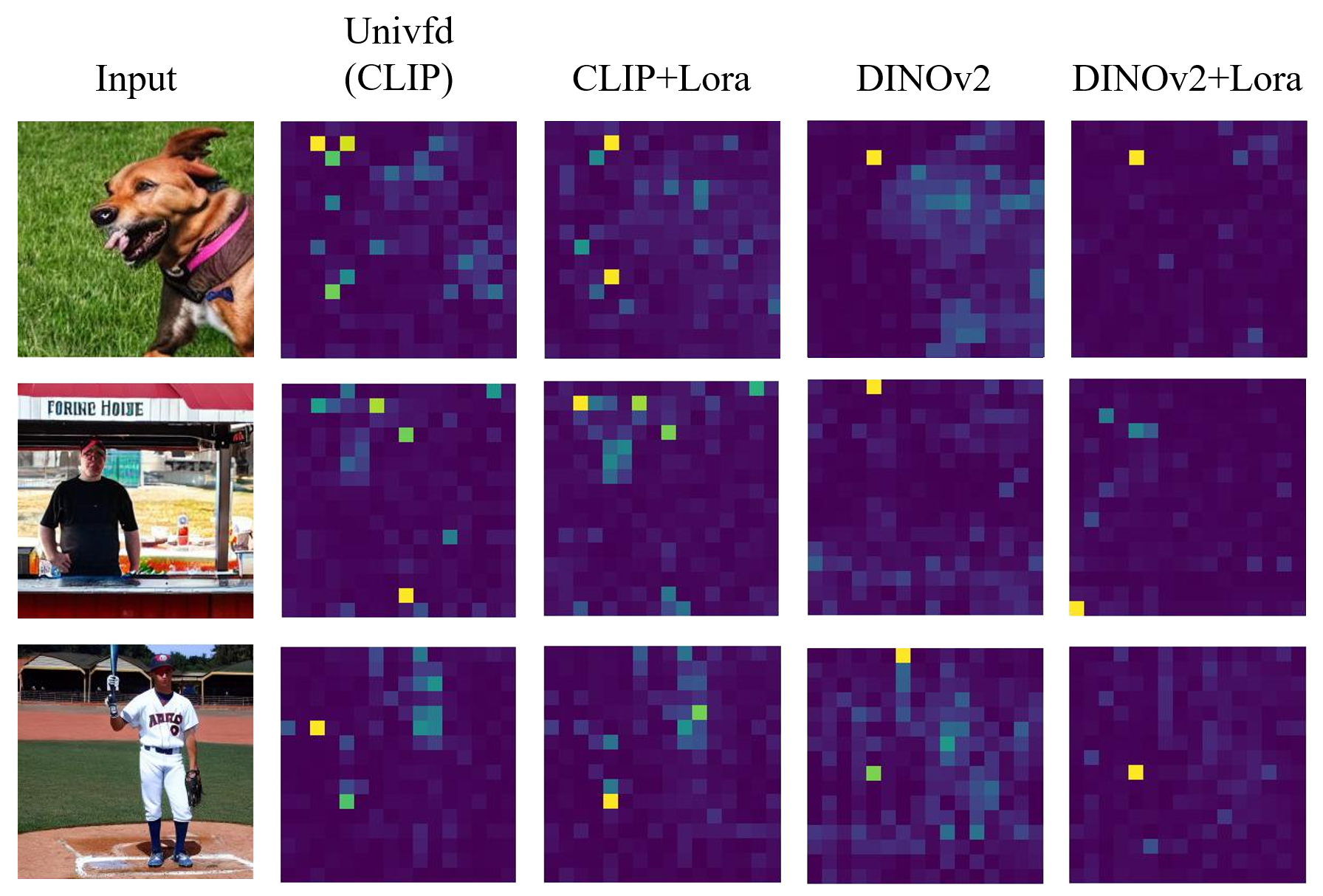}}
\subfloat[Impact of Patch Removal on Recall Rate.]{\includegraphics[height=4.8cm,width=6cm]{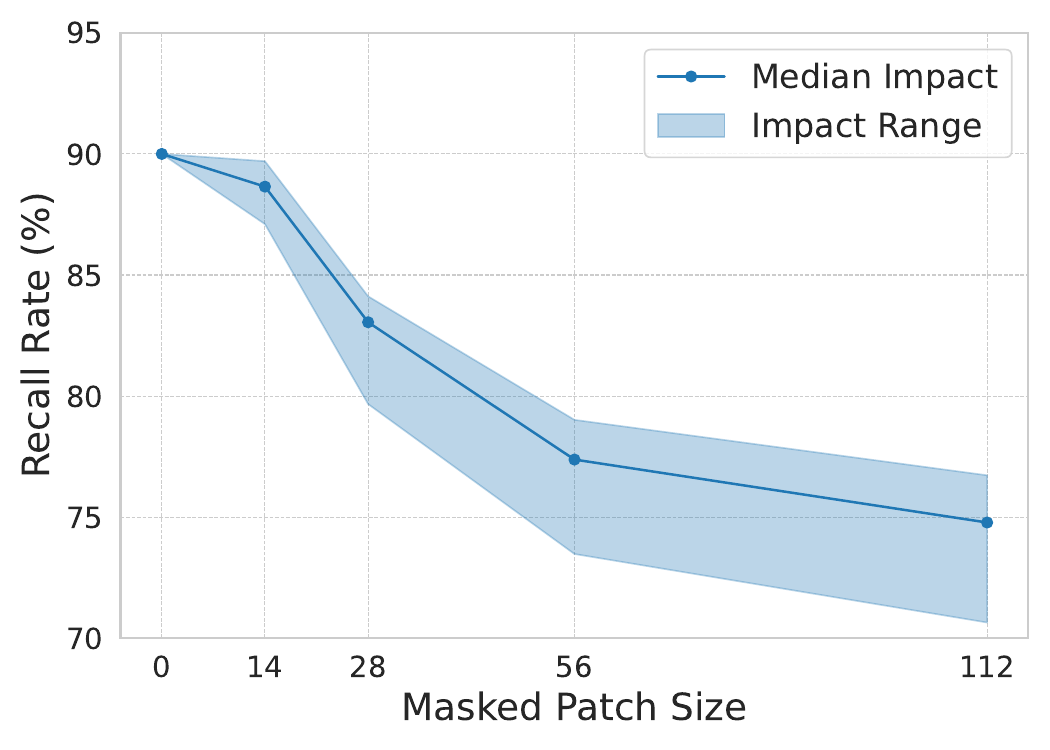}}
\caption{\textbf{(a)} Attention maps reveal the \emph{few-patch bias} of naively trained detectors, where attention concentrates on a small number of dominant patches, reflecting over-reliance on limited regions. \textbf{(b)} Recall degradation occurs when single patches of varying sizes are occluded, showing that detectors are overly sensitive to corruption in specific regions and suffer notable performance drops.}
\label{fig:observations}
\end{figure}

\paragraph{Observations.} 
Our empirical observations indicate that existing detectors often overly rely on a limited number of patches. 
Our experiments reveal that existing detectors tend to over-rely on a limited set of patches.  
Fig.~\ref{fig:observations}(a) shows attention maps from naively trained ViTs, where attention weights concentrate on only a few regions. This phenomenon persists even when changing the backbone or applying LoRA, suggesting a model-agnostic bias.  
To further validate this observation, we systematically mask patches of varying sizes and measured the corresponding recall rate degradation.  
Fig.~\ref{fig:observations}(b) illustrates the performance of UnivFD under patch occlusion. Masking a single patch leads to a substantial drop in accuracy, and the impact varies across different patches, confirming that detectors are disproportionately sensitive to specific regions.

% \begin{wrapfigure}[20]{r}{0.45\textwidth}
%     \centering
%     \vspace{-14pt}
%     \includegraphics[width=0.45\textwidth]{fig/f_acc_vs_masking_size.pdf}
%     \vspace{-12pt}
%     \caption{The illustration of recall degradation in a natively trained model is demonstrated by occluding a single patch of varying sizes. Moreover, a naively trained model may be sensitive to certain patches and may lead to higher drop of recall rate.}
%     \label{fig:degradation_dropout}
% \end{wrapfigure}

% \begin{figure}[t!]
%     \centering
%     \begin{adjustbox}{width=0.95\linewidth}
%     \includegraphics{fig/univfd_and_drct_distribution.png}
%     \end{adjustbox}
%     \caption{\added{CDE} distribution of existing methods.(I will add naive train dino)}
%     \label{fig:dist_tde}
% \end{figure}

\paragraph{Quantitative analysis.} 
Building on the above observations, we employ the \added{Controlled Direct Effect (\added{CDE})} to quantify the impact of each patch.  
Conceptually, if both \(X \rightarrow Y\) and \(Z \rightarrow Y\), then the outcome \(Y\) results from the combined influence of \(X\) and \(Z\). The \added{CDE} measures the contribution of \(X\) by comparing outcomes with and without its effect while keeping other factors fixed.

For an image, the \added{CDE} of the patch at row \(i\), column \(j\) is defined as:  
\begin{equation}
    \added{CDE} \coloneq \delta_{I} - \delta_{I-(i,j)}, 
    \quad \delta \coloneq \text{logit}_{synth} - \text{logit}_{real},
\end{equation}  
where \(I\) denotes the original image and \(I-(i,j)\) the image with the \((i,j)\)-th patch masked (implemented by setting the patch values to zero). By computing the \added{CDE} for each patch, we quantify its relative contribution to the synthetic classification decision.  

\begin{figure}[t!]
    \centering
    \begin{adjustbox}{width=1.0\linewidth}
    \includegraphics{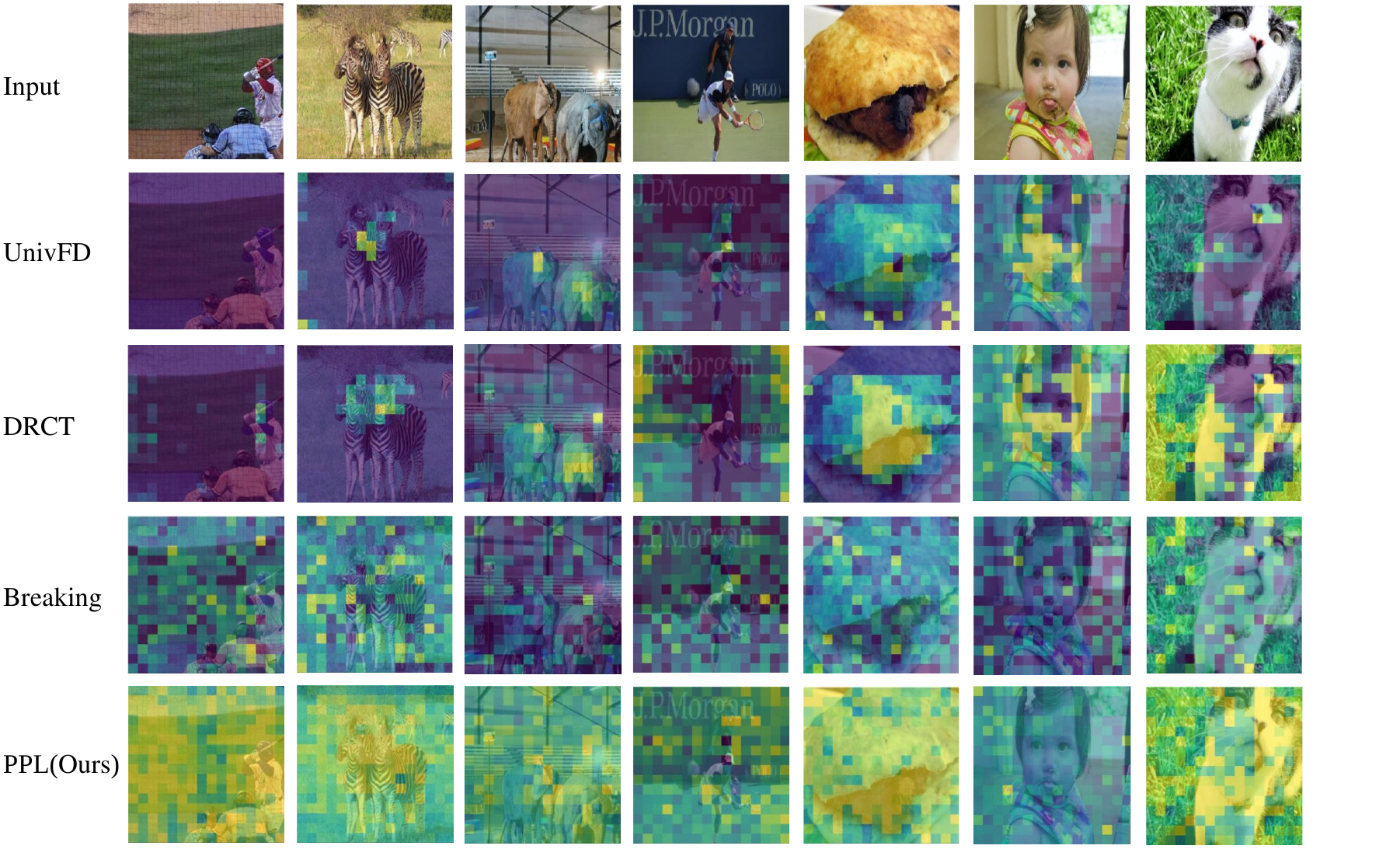}
    \end{adjustbox}
    \caption{\added{CDE} heatmap of existing methods on generated images selected from the DRCT-2M dataset~\citep{chen2024drct}. A broader and more uniform highlighted region indicates a greater number of patches contributing to determining a fake image.}
    \label{fig:visualize_tde}
\end{figure}

Fig.~\ref{fig:visualize_tde} presents \added{CDE} heatmaps. From top to bottom, the number of active patches increases and the \added{CDE} distributions become more uniform. Stronger detectors consistently activate a broader set of patches. These visualizations highlight the prevailing bias toward a few dominant patches with disproportionately high \added{CDE}, motivating us to mitigate few-patch reliance.

\section{Methodology}
\label{sec:method}

\vspace{-5pt}

\begin{figure*}[t!]
    \centering
    \begin{adjustbox}{width=1.0\linewidth}
    \includegraphics{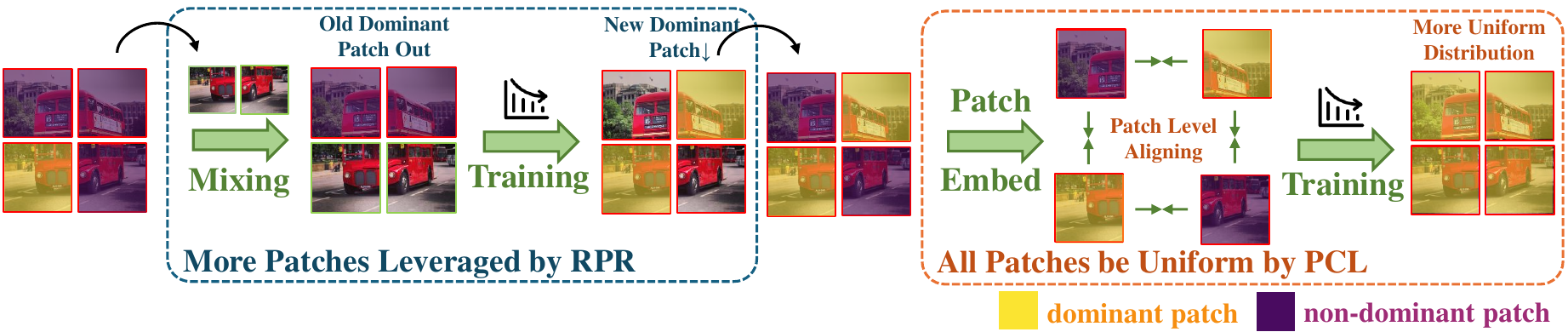}
    \end{adjustbox}
   \caption{The Panoptic Patch Learning (PPL) framework embodies the principles of \textbf{All Patches Matter} and \textbf{More Patches Better} through two key components: Randomized Patch Reconstruction (RPR) and Patch-wise Contrastive Learning (PCL). During training, the model may excessively rely on \textcolor{BurntOrange}{\textbf{dominant patches}}, neglecting others. RPR mitigates this by randomly replacing dominant patches with real ones, encouraging the model to detect artifacts in \textcolor{RedViolet}{\textbf{non-dominant patches}} and thereby expanding the coverage of dominant regions. PCL further promotes balanced patch utilization by aligning the embeddings of patches with the same labels. Together, RPR and PCL foster comprehensive and uniform exploitation of patches.}
    \label{fig:pipeline}
\end{figure*}

Panoptic Patch Learning is a comprehensive framework based on the principles of ``All Patches Matter" and ``More Patches Better," achieved through innovative data and learning strategies, as illustrated in Fig.~\ref{fig:pipeline}. Specifically, the data strategy, Randomized Patch Reconstruction (RPR), discourages the model from over-relying on any specific patches, thereby enhancing its recognition capability for various artifacts across \emph{more patches}. Following this, the learning strategy, Patch-wise Contrastive Learning (PCL), ensures that \emph{all patches}, both frequently attended and underutilized, are brought closer in the feature space, thereby uniformizing the impact of all patches.

\paragraph{Randomized Patch Reconstruction encourages ``More Patches Better".}
The RPR process is carried out by performing diffusion reconstruction on randomly selected patches with a specified proportion, injecting synthetic cues into specific regions of the image while maintaining the overall semantics of the image (as the reconstructed image closely resembles the original image). In practice, RPR is implemented by first applying diffusion reconstruction to the entire image to obtain a reconstructed version. Then, the selected patches in the original image are replaced with their reconstructed counterparts, resulting in a synthetic image where only specific regions contain synthetic artifacts. Here, we emphasize that we inject synthetic features via diffusion reconstruction rather than stitching a synthetic patch, in order to preserve the global semantics and integration of the produced image, and to prevent the model from overfitting to images with disconnected semantics. We use $r \in [0, 1.0]$ to denote the ratio of reconstructed patches relative to the whole image.

% to paired images in which each reconstructed image $I'$  has a corresponding ground truth $I$.
% The images are partitioned into patches, and the patch reconstruction function $\mathcal{R}$ is defined as:
% \begin{equation}
%     \mathcal{R}(P_{i,j}(I')) = 
%     \begin{cases} 
%         P_{i,j}(I) & \text{if } M_{i,j} = 1, \\
%         P_{i,j}(I') & \text{otherwise}
%     \end{cases}
% \end{equation}
% where $M \in \{0,1\}^{m \times m}$ is a random sampled binary mask with reconstruction ratio $r \in [0,1]$.
% When dominant patches are replaced with real patches during this process, training the model on these mixed images forces it to learn artifacts from previously non-dominant patches. 
% By dynamically altering the spatial distribution of attended patches through RPR, the model is forced to learn latent representations from previously underutilized regions, thereby reducing over-reliance on dominant local features. 
% As a result, RPR effectively expands the effective region of the model and enhances its overall performance and robustness.

%TODO:
\paragraph{Patch-wise Contrastive Learning emphasizes ``All Patches Matter".}
PCL operationalizes the principle of ``All Patches Matter" by aligning the embedding vectors of different patches, bringing patches with identical labels closer together while distancing those with different labels. We employ contrastive learning to cluster synthetic patches more closely within each batch, while maintaining a margin that separates synthetic and real patches. This approach ensures that if an image contains any dominant patch with easily learnable artifacts, the model improves its performance on the remaining patches, thus promoting the utilization of all patches. Specifically, for each batch, we utilize a margin-based contrastive loss ~\citep{hadsell2006dimensionality}:

\begin{equation}
\begin{aligned}
\label{eq:contrastive}
& \mathcal{L}_{con}=\sum_{\substack{i,j: \; i \ne j}} \left[ Y \cdot d^2 + (1-Y) \cdot \max\left( 0, \alpha - d^2 \right) \right],
\end{aligned}
\end{equation}
where $i, j$ represent the indices of patch tokens within a batch.
$d$ measures the Euclidean distance between the patch embeddings.
$\alpha$ defines a minimum distance threshold between negative sample pairs, thereby enhancing the model’s ability to distinguish between similar and dissimilar pairs.
$ Y $ indicates whether two patches share identical labels, thus pulling positive patch pairs closer and pushing negative patch pairs further apart.
The overall loss function is a weighted combination of the cross-entropy loss and the patch-wise contrastive loss:
\begin{equation}
\mathcal{L}_{\text{total}} = \lambda \mathcal{L}_{\text{con}} + (1-\lambda) \mathcal{L}_{\text{ce}},
\end{equation}
The practical implementation of PCL is shown in Alg.~\ref{alg:ppl}.

\begin{algorithm}[t!]
\caption{Patch-wise Contrastive Learning (PCL) Training Procedure}
\textbf{Input:} \\
\hspace*{1em} $X$ \quad -- input image tensor after RPR, shape $[B, C, H, W]$ \\
\hspace*{1em} $label_{gt}$ \quad -- image-level ground-truth labels, shape $[B]$ \\
\hspace*{1em} $patch_{gt}$ \quad -- patch-level ground-truth labels, shape $[B, K]$ \\
\hspace*{1em} $\lambda$ \quad -- weighting coefficient for contrastive loss

\begin{algorithmic}[1]
\State $img_{embedding}$, $patch_{embedding}$ $\leftarrow$ $ViT_{Encoder}$($X$) \Comment{$img_{embedding}$: $[B, 1, D]$, $patch_{embedding}$: $[B, K, D]$}
\State $y_{pred}$ $\leftarrow$ Linear($img_{embedding}$) \Comment{Image-level class logits, shape $[B, 2]$}
\State $L_{ce}$ $\leftarrow$ BCELoss($y_{pred}$, $label_{gt}$) \Comment{Image-level classification loss}
\State $L_{con}$ $\leftarrow$ ContrastiveLoss($patch_{embedding}$, $patch_{gt}$) \Comment{Patch-level contrastive loss}
\added{\State $L_{total}$ $\leftarrow$ $\lambda \cdot L_{con} + (1 - \lambda) \cdot L_{ce}$}
\State $L_{total}$.backward()
\end{algorithmic}
\label{alg:ppl}
\end{algorithm}

\section{Experiments}
\label{sec:exper}

% \paragraph{Settings.}
% We compare PPL to other methods across two train-test settings on three datasets: 
% (1) \textbf{Setting-I}: In this setting, the model is trained using real images and images from a single type of generative model. Then the models are evaluated on images from various unseen generative models. This setting assesses the detector's cross-generator generalization ability. The datasets used in Setting-I include GenImage~\citep{zhu2024genimage} and DRCT~\citep{chen2024drct}.

% (2) \textbf{Setting-II}: In this setting, the model has access to a wide range of generative models during the training phase. Then the models are evaluated on a comprehensive dataset that includes challenging cases from modern generative models. Setting-II was proposed by~\citep{yan2024sanity} with the Chameleon dataset.

\paragraph{Implementation details.}
We adopt CLIP~\citep{CLIP} and DINOv2~\citep{oquab2023dinov2} two vision foundation model as backbones , and fine-tune them using LoRA. Unless otherwise specified, in our proposed Panoptic Patch Learning (PPL), image reconstruction is performed with SDv1.4 inpainting at a generation strength of $s=0.25$. The inpainting pipeline uses $step=50$ and guidance scale $7.5$. During training, images are randomly cropped to 224 $\times$ 224, while at test time they are center-cropped to the same resolution.
For the randomized patch reconstruction module, the reconstruction patch size is set to 14 $\times$ 14, consistent with the patch size of ViT. Each fake image in the original training set has a probability of $p_{rpr}=0.9$ of being replaced with a RPR image, where $r_{rpr}=50 \%$ of patches from a real image are randomly selected to do diffusion reconstruction.
For patch-wise contrastive learning, the weight of the contrastive loss is set to $\lambda=0.3$, with a margin parameter $\alpha=1.0$.

\paragraph{Peer methods.} The compared methods involve ResNet-50~\citep{he2016deep}, Conv-B~\citep{liu2022convnet}, Swin-T~\citep{liu2021swin}, CNNSpot~\citep{wang2020cnn}, F3Net~\citep{li2021frequency}, \added{SAFE~\citep{li2024improving}}, UnivFD~\citep{ojha2023towards},  FatFormer~\citep{liu2024forgery}, DRCT~\citep{chen2024drct}, C2P-CLIP~\citep{tan2024c2p}, Effort~\citep{yan2024effort}.

\begin{table*}[t!]
\centering
\caption{Cross-model accuracy (Acc) on GenImage. All methods are trained on the SDv1.4 subset. Results are taken from C2P-CLIP~\citep{tan2024c2p}, except SAFE and Effort, which are reported in their original papers. For Breaking~\citep{zheng2024break}, we re-implement the method because no GenImage results or checkpoints are publicly available. Our results are bolded when they achieve the highest accuracy among all methods.}
\resizebox{1.0\textwidth}{!}{
    \begin{tabular}{l c c c c c c c c c | c}
    \toprule
        Method & Ref & Midjourney & SDv1.4 & SDv1.5 & ADM & GLIDE & Wukong & VQDM & BigGAN & mAcc\\
          \midrule
ResNet-50~\citep{he2016deep}  &  CVPR2016   & 54.9 & 99.9 & 99.7 & 53.5 & 61.9 & 98.2 & 56.6 & 52.0 & 72.1 $\pm$ 22.6 \\
DeiT-S~\citep{touvron2021training}& ICML2021 & 55.6 & 99.9 & 99.8 & 49.8 & 58.1 & 98.9 & 56.9 & 53.5 & 71.6 $\pm$ 23.2\\
Swin-T~\citep{liu2021swin}    &  ICCV2021   & 62.1 & 99.9 & 99.8 & 49.8 & 67.6 & 99.1 & 62.3 & 57.6 & 74.8 $\pm$ 21.1\\
CNNSpot~\citep{wang2020cnn}    &  CVPR2020  & 52.8 & 96.3 & 95.9 & 50.1 & 39.8 & 78.6 & 53.4 & 46.8 & 64.2 $\pm$ 22.6\\
Spec~\citep{zhang2019detecting} & WIFS2019  & 52.0 & 99.4 & 99.2 & 49.7 & 49.8 & 94.8 & 55.6 & 49.8 & 68.8 $\pm$ 24.1\\
F3Net~\citep{qian2020thinking}  & ECCV2020  & 50.1 & 99.9 & 99.9 & 49.9 & 50.0 & 99.9 & 49.9 & 49.9 & 68.7 $\pm$ 25.8\\
GramNet~\citep{liu2020global}   & CVPR2020  & 54.2 & 99.2 & 99.1 & 50.3 & 54.6 & 98.9 & 50.8 & 51.7 & 69.9 $\pm$ 24.2\\
% DIRE       & 60.2 & 99.9 & 99.8 & 50.9 & 55.0 & 99.2 & 50.1 & 50.2 & 70.7 \\
UnivFD~\citep{ojha2023towards}  & CVPR2023  & 93.9 & 96.4 & 96.2 & 71.9 & 85.4 & 94.3 & 81.6 & 90.5 & 88.8 $\pm$ 8.6\\
NPR~\citep{tan2024rethinking}  &  CVPR2024 & 81.0 & 98.2 & 97.9 & 76.9 & 89.8 & 96.9 & 84.1 & 84.2 & 88.6 $\pm$ 8.3\\
FreqNet~\citep{tan2024frequency} & AAAI2024 & 89.6 & 98.8 & 98.6 & 66.8 & 86.5 & 97.3 & 75.8 & 81.4 & 86.8 $\pm$ 11.6\\
% PatchCraft & 79.0 & 89.5 & 89.3 & 77.3 & 78.4 & 89.3 & 83.7 & 72.4 & 82.3 \\
FatFormer~\citep{liu2024forgery} & CVPR2024 & 92.7 & 100.0 & 99.9 & 75.9 & 88.0 & 99.9 & 98.8 & 55.8 & 88.9 $\pm$ 15.7\\
DRCT~\citep{chen2024drct} & ICML2024 & 91.5 & 95.0 & 94.4 & 79.4 & 89.1 & 94.6 & 90.0 & 81.6 & 89.4 $\pm$ 5.9\\
Effort~\citep{yan2024effort} & ICML2025 & 82.4 & 99.8 & 99.8 & 78.7 & 93.3 & 97.4 & 91.7 & 77.6 & 91.1 $\pm$ 11.8\\
Breaking~\citep{zheng2024break} & NIPS2024 & 83.9 & 98.9 & 93.0 & 99.1 & 97.7 & 85.4 & 92.7 & 90.5 & 92.7 $\pm$ 5.8\\
SAFE~\citep{li2024improving} & KDD2025 & 95.3 & 99.4 & 99.3 & 82.1 & 96.3 & 98.2 & 96.3 & 97.8 & 95.6 $\pm$ 5.6\\
C2P-CLIP~\citep{tan2024c2p} & AAAI2025 & 88.2 & 90.9 & 97.9 & 96.4 & 99.0 & 98.8 & 96.5 & 98.7 & 95.8 $\pm$ 4.0\\
\hline
\rowcolor{light-gray} Ours/DINOv2 & & 90.4 & 98.2 & 97.7 & 91.8 & 96.3 & 98.0 & 97.7 & 96.2 & \textbf{95.9} $\pm$ 3.0\\
\rowcolor{light-gray} Ours/CLIP & & 94.8 & 98.5 & 98.3 & 94.7 & 96.1 & 98.6 & 98.5 & 98.0 & \textbf{97.2} $\pm$ 1.7 \\
\bottomrule
    \end{tabular}
  }
\label{tab:SOTA1}
\end{table*}

\begin{table*}[t!]
\centering
\caption{Cross-model accuracy (Acc) on DRCT-2M. All methods are trained on the SDv1.4 subset. Results of other methods are taken from DRCT~\citep{chen2024drct}.} \vspace{-5pt}
\resizebox{1.0\textwidth}{!}{
\begin{tabular}{lcccccccccccccccc|c}
    \toprule
\multirow{2}{*}{Method} & \multicolumn{6}{c}{SD Variants} & \multicolumn{2}{c}{Turbo Variants} & \multicolumn{2}{c}{LCM Variants} & \multicolumn{3}{c}{ControlNet Variants} & \multicolumn{3}{c|}{DR Variants} & \multirow{2}{*}{mAcc}\\ \cline{2-7} \cline{8-9} \cline{10-11} \cline{12-14} \cline{15-17}
 & \thead{LDM} & \thead{SDv1.4} & \thead{SDv1.5} & \thead{SDv2} & \thead{SDXL} & \thead{SDXL-\\Refiner} & \thead{SD-\\Turbo} & \thead{SDXL-\\Turbo} & \thead{LCM-\\SDv1.5} & \thead{LCM-\\SDXL} & \thead{SDv1-\\Ctrl} & \thead{SDv2-\\Ctrl} & \thead{SDXL-\\Ctrl} & \thead{SDv1-\\DR} & \thead{SDv2-\\DR} & \thead{SDX-L\\DR}\\
    \midrule
CNNSpot~\citep{wang2020cnn} & 99.87 & 99.91 & 99.90 & 97.55 & 66.25 & 86.55 & 86.15 & 72.42 & 98.26 & 61.72 & 97.96 & 85.89 & 82.84 & 60.93 & 51.41 & 50.28 & 81.12 $\pm$ 17.6\\
F3Net~\citep{qian2020thinking} & 99.85 & 99.78 & 99.79 & 88.66 & 55.85 & 87.37 & 68.29 & 63.66 & 97.39 & 54.98 & 97.98 & 72.39 & 81.99 & 65.42 & 50.39 & 50.27 & 77.13 $\pm$ 18.1\\
CLIP/RN50~\citep{CLIP} & 99.00 & 99.99 & 99.96 & 94.61 & 62.08 & 91.43 & 83.57 & 64.40 & 98.97 & 57.43 & 99.74 & 80.69 & 82.03 & 65.83 & 50.67 & 50.47 & 80.05 $\pm$ 18.3\\
GramNet~\citep{liu2020global} & 99.40 & 99.01 & 98.84 & 95.30 & 62.63 & 80.68 & 71.19 & 69.32 & 93.05 & 57.02 & 89.97 & 75.55 & 82.68 & 51.23 & 50.01 & 50.08 & 76.62 $\pm$ 17.0\\
De-fake~\citep{sha2023fake} & 92.10 & 99.53 & 99.51 & 89.65 & 64.02 & 69.24 & 92.00 & 93.93 & 99.13 & 70.89 & 58.98 & 62.34 & 66.66 & 50.12 & 50.16 & 50.00 & 75.52 $\pm$ 18.4\\
Conv-B~\citep{liu2022convnet} & 99.97 & 100.0 & 99.97 & 95.84 & 64.44 & 82.00 & 80.82 & 60.75 & 99.27 & 62.33 & 99.80 & 83.40 & 73.28 & 61.65 & 51.79 & 50.41 & 79.11 $\pm$ 18.3\\
UnivFD~\citep{ojha2023towards} & 98.30 & 96.22 & 96.33 & 93.83 & 91.01 & 93.91 & 86.38 & 85.92 & 90.44 & 88.99 & 90.41 & 81.06 & 89.06 & 51.96 & 51.03 & 50.46 & 83.46 $\pm$ 17.0\\
% DIRE & SDv1 & 98.19 & 99.94 & 99.96 & 68.16 & 53.84 & 71.93 & 58.87 & 54.35 & 99.78 & 59.73 & 99.65 & 64.20 & 59.13 & 51.99 & 50.04 & 49.97 & 71.23 \\
% DIRE & SDv2 & 54.62 & 75.89 & 76.04 & 99.87 & 59.90 & 93.08 & 99.77 & 57.55 & 87.29 & 72.53 & 67.85 & 99.69 & 64.40 & 49.96 & 52.48 & 49.92 & 72.55 \\
% DRCT/Conv-B & SDv1 & 99.91 & 99.90 & 99.90 & 96.32 & 83.87 & 85.63 & 91.88 & 70.04 & 99.66 & 78.76 & 99.90 & 95.01 & 81.21 & 99.90 & 95.40 & 75.39 & 90.79 \\
% DRCT/Conv-B & SDv2 & 99.66 & 98.56 & 98.48 & 99.85 & 96.10 & 98.68 & 99.59 & 83.30 & 98.45 & 93.78 & 96.68 & 99.85 & 97.66 & 93.91 & 99.87 & 90.39 & 96.55 \\
 %DRCT/UnivFD & SDv1 & 96.74 & 96.26 & 96.33 & 94.89 & 96.24 & 93.46 & 93.43 & 92.94 & 91.17 & 95.01 & 95.60 & 92.68 & 91.95 & 94.10 & 69.55 & 57.43 & 90.49 \\
DRCT~\citep{chen2024drct} & 94.45 & 94.35 & 94.24 & 95.05 & 95.61 & 95.38 & 94.81 & 94.48 & 91.66 & 95.54 & 93.86 & 93.48 & 93.54 & 84.34 & 83.20 & 67.61 & 91.35 $\pm$ 4.7\\
\hline
% \rowcolor{light-gray} Ours/DINOv2 & 99.82 & 99.82 & 99.82 & 99.55 & 99.55 & 86.73 & 98.76 & 96.19 & 99.51 & 99.67 & 99.82 & 99.61 & 97.69 & 99.82 & 97.68 & 77.33 & 96.96 \\
\rowcolor{light-gray} Ours/DINOv2 & 99.55 & 99.55 & 99.55 & 99.54 & 99.55 & 94.70 & 99.53 & 99.23 & 99.31 & 99.55 & 99.54 & 99.55 & 99.39 & 99.48 & 99.55 & 97.42 & \underline{99.06} $\pm$ 0.1\\
% \rowcolor{light-gray} Ours/CLIP & 99.65 & 99.60 & 99.61 & 99.03 & 98.92 & 90.87 & 97.50 & 94.69 & 98.67 & 98.35 & 99.62 & 97.61 & 93.36 & 99.63 & 95.78 & 68.56 & 95.72 \\
\rowcolor{light-gray} Ours/CLIP & 99.70 & 99.70 & 99.69 & 99.67 & 99.71 & 99.40 & 99.48 & 99.40 & 99.62 & 99.70 & 99.68 & 99.64 & 99.51 & 99.61 & 99.67 & 97.80 & \textbf{99.50} $\pm$ 0.1\\
\bottomrule
    \end{tabular}
}  \vspace{-5pt}
  \label{tab:SOTA2}
\end{table*}

\subsection{Comparison with other methods} 

\paragraph{Comparison on GenImage.}
Tab.~\ref{tab:SOTA1} compares PPL with other methods on GenImage. We observe: (1) PPL consistently achieves higher accuracy across different backbones. (2) The standard deviation of PPL’s accuracy is smaller, indicating improved stability in detecting diverse generative models.

\paragraph{Comparison on DRCT-2M.}
Tab.~\ref{tab:SOTA2} reports results on DRCT-2M. The results indicate: (1) PPL consistently achieves SoTA with the lowest std, demonstrating both effectiveness and stability. (2) While DRCT shows relatively poor performance on SDXL-related subsets, PPL maintains a more balanced performance across diverse subsets, underscoring its robustness.

% \paragraph{Cross-Dataset Comparison on Benchmarks Including GANs} 
% To evaluate the generalizability of our approach beyond diffusion models, we tested our best-performing model—trained on the SDv1.4 subset of the GenImage dataset—on two benchmarks that include GAN-generated images.

\paragraph{Comparison on AIGCDetectBenchmark and UniversalFakeDetect.}
Tab.~\ref{tab:SOTA5} and Tab.~\ref{tab:SOTA6} present results on AIGCDetectBenchmark~\citep{zhong2024patchcraft} and UniversalFakeDetect~\citep{ojha2023towards}, respectively. The results of baseline methods are taken from~\citep{yan2024sanity} and ~\citep{tan2024c2p}. We observe that PPL, when trained solely on diffusion-generated data, generalizes effectively to detecting GAN-generated images—even surpassing baseline methods trained directly on GAN data—highlighting PPL’s strong generalization capability.

\paragraph{Comparison on the in-the-wild dataset Chameleon.}
Tab.~\ref{tab:SOTA3} reports results on Chameleon, a challenging dataset comprising diverse images collected from online websites. We observe that most existing methods only marginally exceed the accuracy of random guessing (50\%). In contrast, PPL achieves 70\% accuracy on Chameleon, demonstrating strong generalization on real-world data.

\begin{table*}[!t]
    \centering
    \caption{Cross-dataset and cross-model accuracy (mAcc) on AIGCDetectionBenchmark. PPL is trained on the GenImage SDv1.4 subset due to its reliance on diffusion-based reconstruction. Baseline methods are trained on ProGAN data provided by AIGCDetectionBenchmark, which is more in-distribution with the test set, thereby giving them an inherent advantage under this setting. Baseline results are taken from AIDE~\citep{yan2024sanity}.} \vspace{-5pt}
  \label{tab:SOTA5}
\adjustbox{width=\textwidth}{
    \begin{tabular}{lcccccccccccccccc|cl}
\toprule
Method & ProGAN & StyleGAN & BigGAN & CycleGAN & StarGAN & GauGAN & StyleGAN2 & WFR & ADM & Glide & Midjourney & SD v1.4 & SD v1.5 & VQDM & Wukong & DALLE2 & mAcc \\
\midrule
CNNSpot & 100.00 & 90.17 & 71.17 & 87.62 & 94.60 & 81.42 & 86.91 & 91.65 & 60.39 & 58.07 & 51.39 & 50.57 & 50.53 & 56.46 & 51.03 & 50.45 & 70.78 $\pm$ 18.30\\
FreDect & 99.36 & 78.02 & 81.97 & 78.77 & 94.62 & 80.57 & 66.19 & 50.75 & 63.42 & 54.13 & 45.87 & 38.79 & 39.21 & 77.80 & 40.30 & 34.70 & 64.03 $\pm$ 20.41\\
Fusing & 100.00 & 85.20 & 77.40 & 87.00 & 97.00 & 77.00 & 83.30 & 66.80 & 49.00 & 57.20 & 52.20 & 51.00 & 51.40 & 55.10 & 51.70 & 52.80 & 68.38 $\pm$ 17.46\\
LNP & 99.67 & 91.75 & 77.75 & 84.10 & 99.92 & 75.39 & 94.64 & 70.85 & 84.73 & 80.52 & 65.55 & 85.55 & 85.67 & 74.46 & 82.06 & 88.75 & 83.84 $\pm$ 9.46\\
LGrad & 99.83 & 91.08 & 85.62 & 86.94 & 99.27 & 78.46 & 85.32 & 55.70 & 67.15 & 66.11 & 65.35 & 63.02 & 63.67 & 72.99 & 59.55 & 65.45 & 75.34 $\pm$ 13.8\\
UnivFD & 99.81 & 84.93 & 95.08 & 98.33 & 95.75 & 99.47 & 74.96 & 86.90 & 66.87 & 62.46 & 56.13 & 63.66 & 63.49 & 85.31 & 70.93 & 50.75 & 78.43 $\pm$ 16.19\\
DIRE-G & 95.19 & 83.03 & 70.12 & 74.19 & 95.47 & 67.79 & 75.31 & 58.05 & 75.78 & 71.75 & 58.01 & 49.74 & 49.83 & 53.68 & 54.46 & 66.48 & 68.68 $\pm$ 14.00\\
DIRE-D & 52.75 & 51.31 & 49.70 & 49.58 & 46.72 & 51.23 & 51.72 & 53.30 & 98.25 & 92.42 & 89.45 & 91.24 & 91.63 & 91.90 & 90.90 & 92.45 & 71.53 $\pm$ 20.86\\
PatchCraft & 100.00 & 92.77 & 95.80 & 70.17 & 99.97 & 71.58 & 89.55 & 85.80 & 82.17 & 83.79 & 90.12 & 95.38 & 95.30 & 88.91 & 91.07 & 96.60 & 89.31 $\pm$ 8.61\\
AIDE & 99.99 & 99.64 & 83.95 & 98.48 & 99.91 & 73.25 & 98.00 & 94.20 & 93.43 & 95.09 & 77.20 & 93.01 & 92.85 & 95.16 & 93.55 & 96.60 & 92.77 $\pm$ 7.66\\
NPR & 99.79 & 97.70 & 84.35 & 96.10 & 99.35 & 82.50 & 98.38 & 65.80 & 69.69 & 78.36 & 77.85 & 78.63 & 78.89 & 78.13 & 76.11 & 64.90 & 82.91 $\pm$ 11.54\\
\hline
\rowcolor{light-gray} Ours/DINOv2 & 96.94 & 94.27 & 94.73 & 89.44 & 89.99 & 93.99 & 89.44 & 95.00 & 91.02 & 97.84 & 85.00 & 99.43 & 99.03 & 99.17 & 99.26 & 96.05 & \textbf{94.41} $\pm$ 4.20\\
\rowcolor{light-gray} Ours/CLIP & 89.12 & 89.94 & 83.57 & 97.16 & 97.12 & 75.29 & 89.17 & 95.20 & 94.67 & 96.05 & 94.78 & 98.49 & 98.19 & 98.53 & 98.61 & 97.90 & \textbf{93.36} $\pm$ 6.31\\
\hline
    \end{tabular} 
} \vspace{-10pt}
\end{table*}

\begin{table*}[!t]
    \centering
    \caption{Cross-dataset and cross-model accuracy (mAcc) on the UniversalFakeDetect. PPL is trained on the SDv1.4 subset of GenImage, while other methods are trained on GAN. Baseline results are taken from C2P-CLIP~\citep{tan2024c2p}.} \vspace{-5pt}
  \label{tab:SOTA6}
\adjustbox{width=\textwidth}{
    \begin{tabular}{l c c c c c c c c c c c c c c c c c}
    \bottomrule 
    \hline
      \multirow{3}*{Methods}  & \multirow{3}*{Ref} & \multicolumn{6}{c}{GAN} &  \multirow{3}*{Guided} & \multicolumn{3}{c}{LDM} & \multicolumn{3}{c}{GLIDE} &  \multirow{3}*{DALLE}  &  \multirow{3}*{mAcc}\\ 
      
       \cmidrule(r){3-8} \cmidrule(r){10-12}   \cmidrule(r){13-15}  
       ~ & ~ & \makecell[c]{Pro-\\GAN} & \makecell[c]{Cycle-\\GAN} & \makecell[c]{Big-\\GAN} & \makecell[c]{Style-\\GAN}  & \makecell[c]{Gau-\\GAN}  & \makecell[c]{Star-\\GAN} & ~ & {\makecell[c]{200\\steps}}& {\makecell[c]{200\\w/cfg}}& {\makecell[c]{100\\steps}}& {\makecell[c]{100\\27}} & {\makecell[c]{50\\27}} & \makecell[c]{100\\10} & ~ & ~ & ~\\
       \hline
       CNN-Spot & CVPR2020 & 99.99 & 85.20 & 70.20 & 85.70 & 78.95 & 91.70 & 60.07 & 54.03 & 54.96 & 54.14 & 60.78 & 63.80 & 65.66 & 55.58 & 70.05 $\pm$ 14.90 \\
       Patchfor & ECCV2020 & 75.03 & 68.97 & 68.47 & 79.16 & 64.23 & 63.94 & 67.41 & 76.50 & 76.10 & 75.77 & 74.81 & 73.28 & 68.52 & 67.91 & 71.44 $\pm$ 4.73 \\
       Co-occurence & Elect. Imag. &  97.70 & 63.15 & 53.75 & 92.50 & 51.10 & 54.70 & 60.50 & 70.70 & 70.55 & 71.00 & 70.25 & 69.60 & 69.90 & 67.55 & 68.78 $\pm$ 12.68 \\
       Freq-spec  & WIFS2019 & 49.90 & 99.90 & 50.50 & 49.90 & 50.30 & 99.70 & 50.90 & 50.40 & 50.40 & 50.30 & 51.70 & 51.40 & 50.40 & 50.00 & 57.55 $\pm$ 17.25 \\
       F3Net & ECCV2020 & 99.38 & 76.38 & 65.33 & 92.56 & 58.10 & 100.00 & 69.20 & 68.15 & 75.35 & 68.80 & 81.65 & 83.25 & 83.05 & 66.30 & 77.68 $\pm$ 12.47 \\
       UnivFD & CVPR2023 & 100.00 & 98.50 & 94.50 & 82.00 & 99.50 & 97.00 & 70.03 & 94.19 & 73.76 & 94.36 & 79.07 & 79.85 & 78.14 & 86.78 & 87.69 $\pm$ 9.97 \\
       LGrad & CVPR2023 & 99.84 & 85.39 & 82.88 & 94.83 & 72.45 & 99.62 & 77.50 & 94.20 & 95.85 & 94.80 & 87.40 & 90.70 & 89.55 & 88.35 & 89.53 $\pm$ 7.70 \\
       FreqNet & AAAI2024  & 97.90 & 95.84 & 90.45 & 97.55 & 90.24 & 93.41 & 86.70 & 84.55 & 99.58 & 65.56 & 85.69 & 97.40 & 88.15 & 59.06 & 88.01 $\pm$ 11.56 \\ 
       NPR & CVPR2024 & 99.84 & 95.00 & 87.55 & 96.23 & 86.57 & 99.75 & 84.55 & 97.65 & 98.00 & 98.20 & 96.25 & 97.15 & 97.35 & 87.15 & 94.37 $\pm$ 5.19 \\
       FatFormer & CVPR2024  & 99.89 & 99.32 & 99.50 & 97.15 & 99.41 & 99.75 & 76.00 & 98.60 & 94.90 & 98.65 & 94.35 & 94.65 & 94.20 & 98.75 & \textbf{96.08} $\pm$ 5.95 \\
        C2P-CLIP & AAAI2025 & 99.98 & 97.31 & 99.12 & 96.44 & 99.17 & 99.60 & 69.10 & 99.25 & 97.25 & 99.30 & 95.25 & 95.25 & 96.10 & 98.55 & 95.83 $\pm$ 7.57 \\
       \hline
        \rowcolor{light-gray} Ours/DINOv2 & & 96.94 & 89.44 & 94.73 & 94.27 & 93.99 & 89.99 & 92.30 & 99.40 & 99.40 & 99.40 & 98.35 & 98.60 & 98.50 & 99.40 & 96.05 $\pm$ 3.58 \\
        \rowcolor{light-gray} Ours/CLIP & & 89.12 & 97.16 & 83.57 & 89.94 & 75.29 & 97.12 & 94.55 & 98.80 & 98.80 & 98.80 & 96.90 & 97.40 & 97.65 & 98.80 & 93.85 $\pm$ 6.78 \\
\hline
% \bottomrule
    \end{tabular} 
}  \vspace{-10pt}
\end{table*}

\begin{table*}[t!]
\renewcommand{\arraystretch}{1.5}
\centering
\caption{Cross-dataset accuracy (Acc) on the in-the-wild dataset Chameleon. Results of other methods are directly taken from AIDE~\citep{yan2024sanity}. For each training dataset, the first row reports the overall \textbf{Acc} on the Chameleon test set, while the second row presents \textbf{Acc} separately for \textbf{fake images} and \textbf{real images} for detailed analysis.} \vspace{-5pt}
\resizebox{1.0\textwidth}{!}{
\begin{tabular}{cccccccccccc}
\toprule
\textbf{Training Dataset} & \textbf{CNNSpot} & \textbf{FreDect} & \textbf{Fusing} & \textbf{UnivFD} & \textbf{DIRE} & \textbf{PatchCraft} & \textbf{NPR} & \textbf{AIDE} & \textbf{Ours/CLIP} & \textbf{Ours/DINOv2}\\
\midrule
\multirow{2}{*}{SD v1.4} & 60.11 & 56.86 & 57.07 & 55.62 & 59.71 & 56.32 & 58.13 & 62.60 & \underline{63.94} & \textbf{66.63} \\
 & 8.86/98.63 & 1.37/98.57 & 0.00/99.96 & 17.65/93.50 & 11.86/95.67 & 3.07/96.35 & 2.43/100.00 & 20.33/94.38 & 17.27/99.01 & 64.65/68.12\\
\cline{2-11}
\multirow{2}{*}{All GenImage} & 60.89 & 57.22 & 57.09 & 60.42 & 57.83 & 55.70 & 57.81 & 65.77 & \underline{69.33} & \textbf{72.07} \\
 & 9.86/99.25 & 0.89/99.55 & 0.02/99.98 & 85.52/41.56 & 2.09/99.73 & 1.39/96.52 & 1.68/100.00 & 26.80/95.06 & 38.93/92.18 & 49.68/88.99\\
\bottomrule
\end{tabular} 
} \vspace{-5pt}
\label{tab:SOTA3}
\end{table*}

\subsection{Robustness Studies}
\vspace{-5pt}

\begin{figure}[t!]
    \centering
    \begin{minipage}{\textwidth} % This makes the entire figure take only half the column width
        \centering
        \includegraphics[width=0.99\textwidth]{fig/Robustness/robust_legend.png}
        \vspace{0.2cm}
        \begin{subfigure}[b]{0.32\textwidth}
            \includegraphics[width=\textwidth]{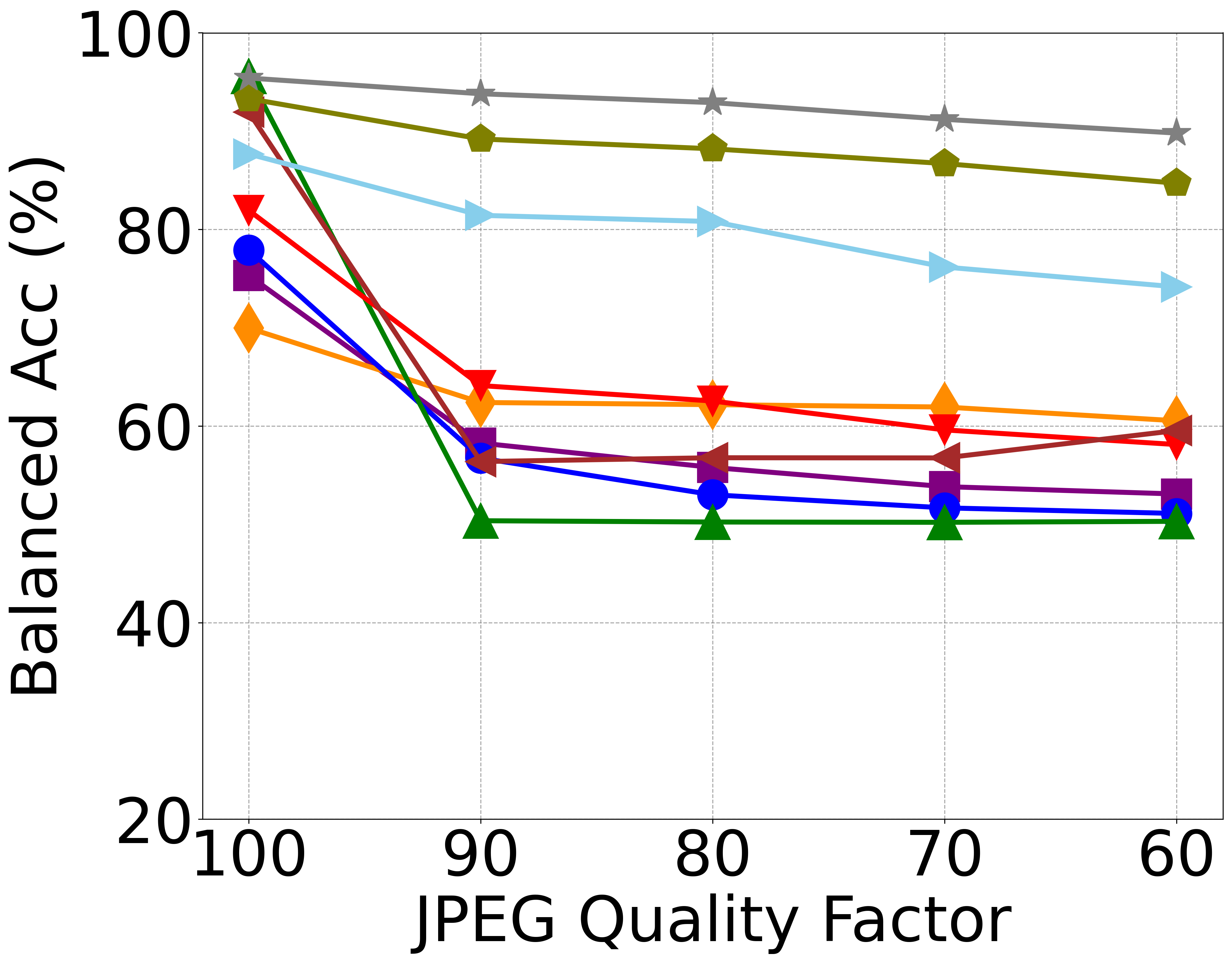}
        \end{subfigure}%
        \hfill%
        \begin{subfigure}[b]{0.32\textwidth}
            \includegraphics[width=\textwidth]{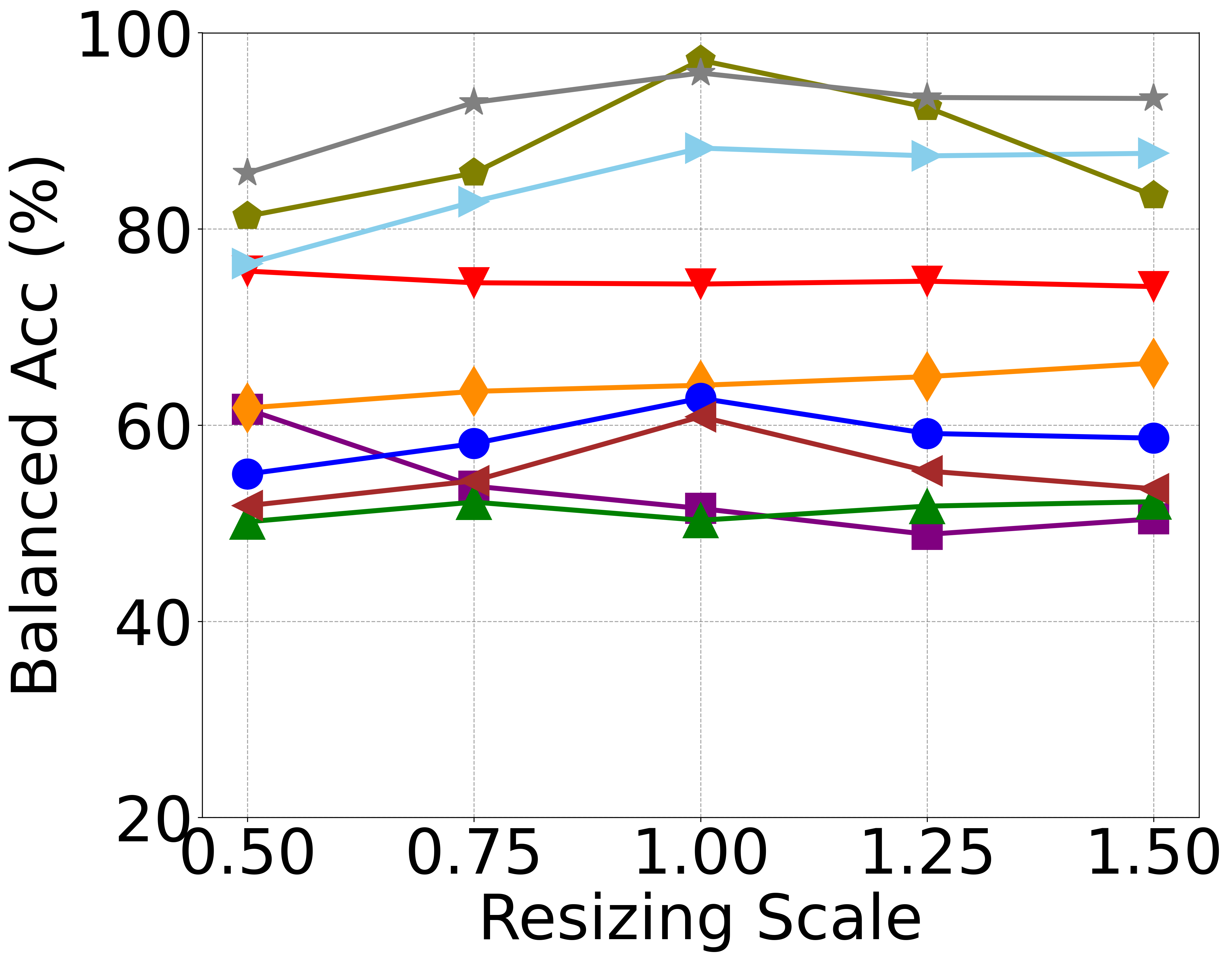}
        \end{subfigure}%
        \hfill%
        \begin{subfigure}[b]{0.32\textwidth}
            \includegraphics[width=\textwidth]{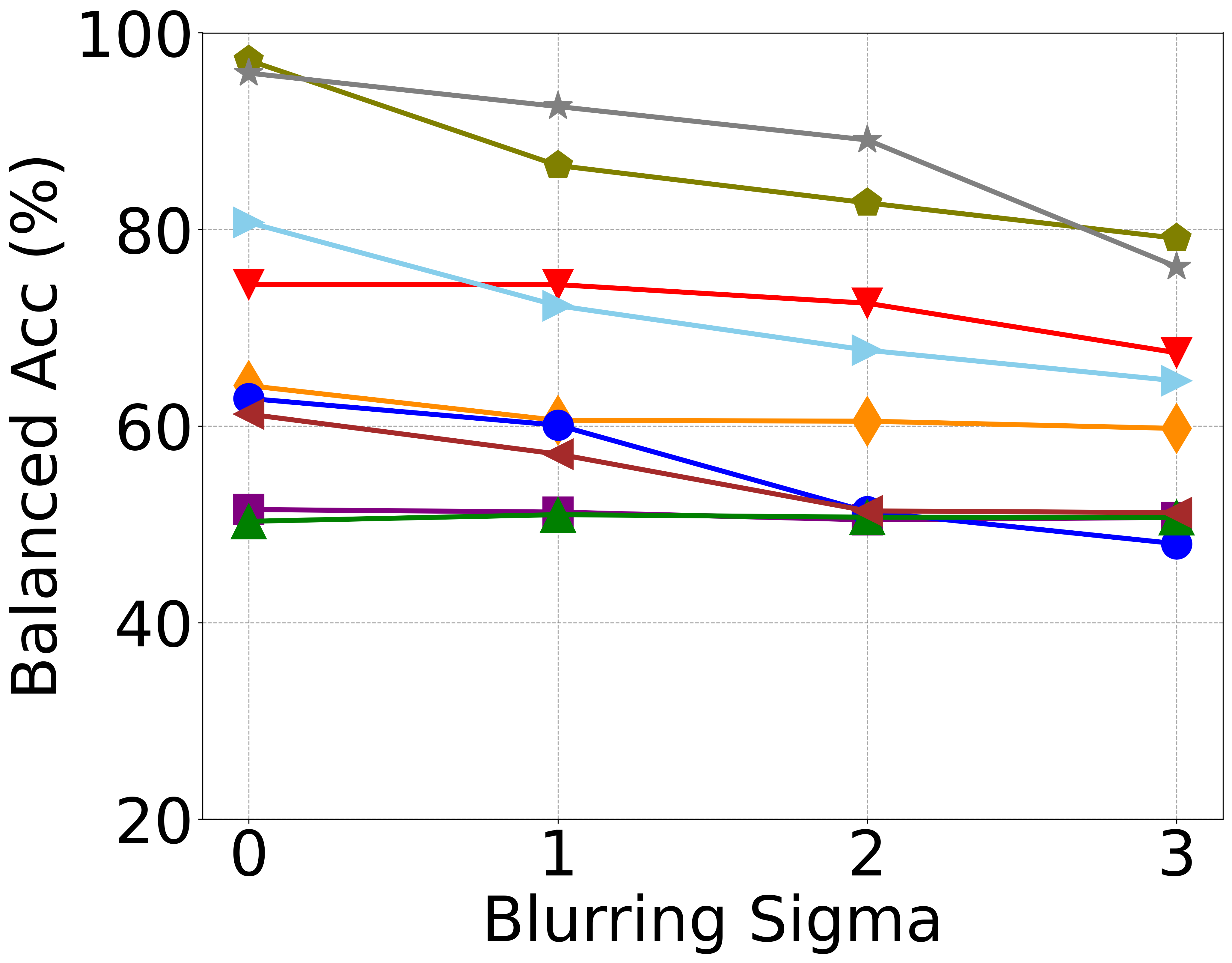}
        \end{subfigure}
    \end{minipage} \vspace{-15pt}
    \caption{Robustness to image corruptions on GenImage Dataset. Performance is evaluated under JPEG compression (quality factor $Q \in \{100, 90, 80, 70, 60\}$), Gaussian blur (standard deviation $\sigma \in \{0.0, 1.0, 2.0, 3.0\}$), and resizing (scaling factor $S \in \{0.5, 0.75, 1.0, 1.25, 1.5\}$).} \vspace{-10pt}
    \label{fig:robustness}
\end{figure}

We conduct robustness experiments on GenImage to evaluate the reliability of our method under common image corruptions. As shown in Fig.~\ref{fig:robustness}, both CLIP and DINOv2 backbones sustain high accuracy even under severe JPEG compression and strong Gaussian blur, demonstrating the robustness of our approach.

\begin{figure}[t!]
    \centering
    \begin{adjustbox}{width=1.0\linewidth}
    \includegraphics{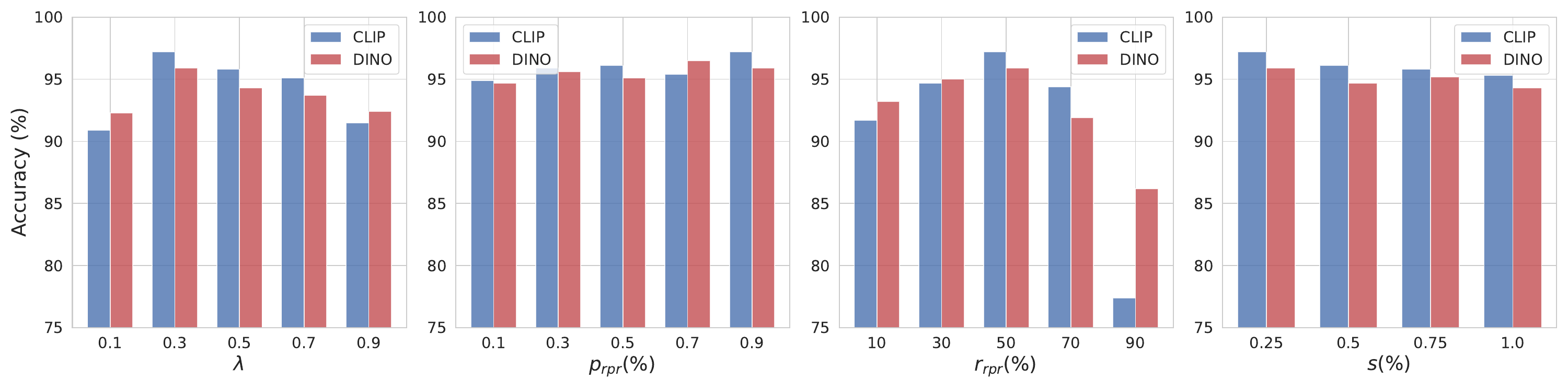}
    \end{adjustbox} \vspace{-20pt}
    \caption{Ablation study on hyperparameters: $\lambda$ representing the weight of contrastive loss. $p_{rpr}$ representing the probability of replacing the fake images with RPR images during training. $r_{rpr}$ representing the ratio of reconstructed patches, and $s$ representing the strength of reconstruction.} \vspace{-15pt}
    \label{fig:hyper}
\end{figure}

\subsection{Ablation Studies}

\paragraph{Ablation on the hyperparameters.}
Fig.~\ref{fig:hyper} reports the influence of key hyperparameters in PPL. The results suggest four main observations:  
(1) PPL achieves peak accuracy at $\lambda = 0.3$.  
(2) PPL is relatively robust to $p_{rpr}$, the probability of replacing fake images with RPR images during training.  
(3) PPL is sensitive to the patch reconstruction ratio $r_{rpr}$, where accuracy degrades at excessively high ratios, with the best performance obtained around $r_{rpr}=50\%$.  
(4) Smaller reconstruction strength $s$ not only improves accuracy but also reduces computational cost, making it preferable in practice.

% (1) The overall accuracy performance is relatively robust to the hyperparemeter setting.
% (2) Generally, accuracy improves as the mixing ratio increases from 10\% to 50\%, peaking at 50\%, but experiences a significant decline at 90\%. This suggests that replacing too many patches in an AI-generated image with real ones sets an overly challenging goal for the model, which hurts its performance.

\begin{wrapfigure}[12]{r}{0.45\textwidth}
    \centering 
    \vspace{-15pt}
    \includegraphics[width=0.45\textwidth]{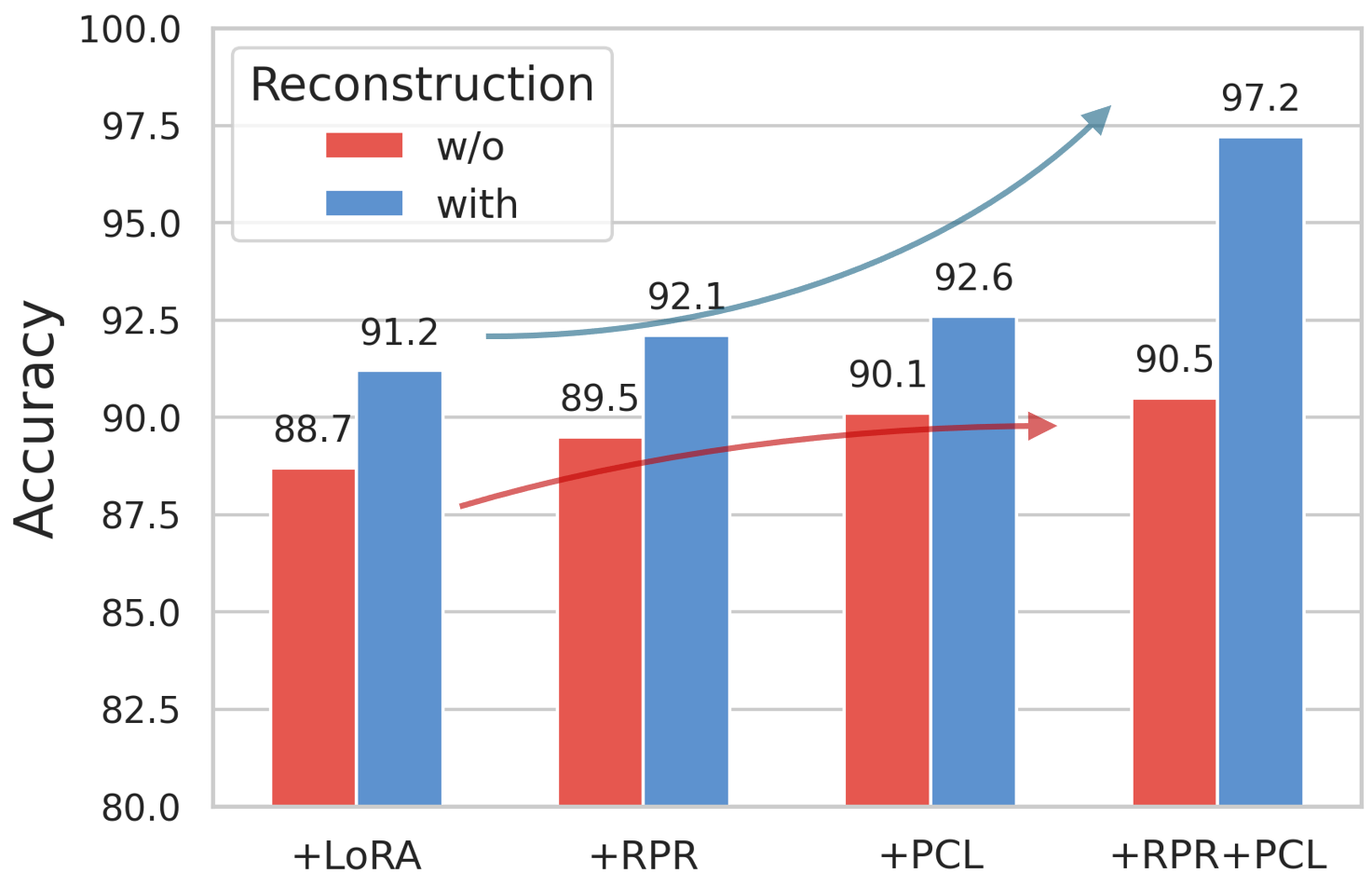}
    \vspace{-15pt}
    \caption{Ablation study on each module.}
    \label{fig:ablation}
\end{wrapfigure}

\paragraph{Ablation on the impact of each module.}
Fig.~\ref{fig:ablation} highlights the contributions of Randomized Patch Reconstruction (RPR) and Patch-wise Contrastive Learning (PCL). We compare two strategies for injecting synthetic artifacts into real images: (1) diffusion-based reconstruction (\textcolor{blue}{blue}), and (2) random patch replacement from \added{original} synthetic images (\textcolor{red}{red}). The results show that reconstruction serves as a more effective mechanism for introducing synthetic cues to guide the model. While either RPR or PCL alone enhances performance, their combination yields markedly stronger improvements. Additional ablation studies are provided in the Appendix due to space constraints.

% Additional ablation studies are provided in the Appendix due to space constraints.

% Naive fine-tuning of CLIP with LoRA produces only marginal gains. Incorporating either RPR or PCL into the LoRA fine-tuning pipeline increases accuracy, demonstrating the effectiveness of each module; the best results occur when RPR and PCL are used together. By contrast, replacing the proposed fake-artifact generation with the initial fake image yields only a negligible improvement.

\vspace{-10pt}
\section{Conclusion}
\label{sec:con}
\vspace{-5pt}

Our work is based on the nature of the AIGI detection problem, which can be concluded as ``All Patches Matter, More Patches Better." 
However, our observations indicate that existing detectors are unable to fully take advantage of all patches in an AI-generated image. To address this issue, we propose a randomized patch reconstruction augmentation combined with patch-wise contrastive learning strategy. 
This approach effectively prevents the model from becoming a lazy learner and enhances the utilization of every patch. We achieve state-of-the-art performance on several well-known academic datasets across various benchmark datasets. 
The outstanding performance achieved in both settings supports our findings and proves the efficacy of the proposed learning framework.

\section*{Acknowledgments}
\label{sec:ack}
\vspace{-5pt}

This work was supported in part by the National Science Foundation for Distinguished Young Scholars under Grant 62225605, ``Pioneer" and ``Leading Goose" R\&D Program of Zhejiang (No.2025C02014), Ningbo Science and Technology Special Projects under Grant No.2025Z028, Project 12326608 supported by NSFC, and the Fundamental Research Funds for the Central Universities.

\bibliography{iclr2026_conference}
\bibliographystyle{iclr2026_conference}

\newpage

\appendix

This appendix provides supplementary details and additional analyses. Section~\ref{sec:implementation} describes the implementation setup; Section~\ref{sec:ablation} reports more ablation studies; and Section~\ref{sec:comparative} presents a comparative \added{CDE} analysis with both statistical and visual demonstrations.

\section{Implementation Details}
\label{sec:implementation}

\paragraph{Implementation of RPR.}
In Randomized Patch Reconstruction (RPR), diffusion-reconstructed real images are replaced with their original counterparts on a patch-wise basis. 
For consistency, we use Stable Diffusion v1 (SDv1) as the reconstruction model and reconstruct real images in the training subset of SDv1.4 from both GenImage and DRCT. 
The reconstruction is performed with the inpainting pipeline (50 denoising steps, guidance scale $7.5$) using an empty prompt, a zero-filled mask of the same size as the input image, and the original image as inputs.

\paragraph{Implementation of PCL.}
Patch-wise Contrastive Learning (PCL) introduces moderate computational overhead. 
Training a LoRA model with only binary cross-entropy (BCE) loss on CLIP-Large requires 16 GB of GPU memory, while incorporating the margin-based contrastive loss increases memory usage to 19 GB. 
Similarly, training one epoch on GenImage with an NVIDIA V100 GPU increases the runtime from 3 to 4 hours.

\section{Additional Ablation Studies}
\label{sec:ablation}
We investigate the impact of different contrastive losses applied to embedded patch tokens. 
Two widely used losses are considered: InfoNCE and the margin-based contrastive loss. 
InfoNCE maximizes similarity between positive pairs while minimizing it for negative pairs, typically formulated as:

\begin{equation}
\mathcal{L}_{\text{q}} = - \log \frac{\exp(q \cdot k_+/\tau)}{\sum_{i=0}^N \exp(q \cdot k_i/\tau)},
\end{equation}

where \(q\) and \(k_+\) denote the embeddings of the sample and its positive counterpart, and \(\tau\) is a temperature parameter. InfoNCE requires computing similarities between each sample and all negative samples, resulting in a computational complexity that grows quadratically with batch size.

In contrast, the margin-based contrastive loss, constrains the Euclidean distance between sample pairs by pulling positive pairs closer and pushing negative pairs apart with a margin \(\alpha\):

\begin{equation}
\mathcal{L}_{contrastive} = \sum_{i,j: \; i \neq j} \left[ Y \cdot d_{ij}^2 + (1 - Y) \cdot \max(0, \alpha - d_{ij}^2) \right],
\end{equation}

where \(d_{ij} = \| \text{Emb}_{\text{pat}}^i - \text{Emb}_{\text{pat}}^j \|_2\) is the Euclidean distance between embedded patch tokens, and \(Y\) is an indicator function that equals 1 if the pair shares the same label and 0 otherwise. This loss only penalizes negative pairs whose distance is less than the margin, thus avoiding computations over all negative pairs and reducing the overall computational cost. The \(d_{ij} \) could also use the cosine distance, which is also compared in our experiments.

\paragraph{Impact of loss function.} Tab.~\ref{tab:choice} illustrates the effectiveness of these loss functions. Unless otherwise specified, all experiments reported in the appendix are conducted on the GenImage dataset, and performance is measured using mean accuracy (mAcc).

\paragraph{Impact of randomized patch reconstruction vs. fixed position reconstruction.}
Our randomized patch reconstruction method employs a random selection process for image reconstruction, allowing fake patches to appear in different regions across the entire image. Alternatively, replacing patches at fixed positions also yields images composed of both real and synthetic elements. Tab.~\ref{tab:mix} illustrates the effectiveness of randomized patch reconstruction compared to fixed-position reconstruction, both of which leverage patch-wise contrastive learning. 

\begin{table*}[t]
    \centering
    \caption{The impact of contrastive loss choice. }
\resizebox{1.0\textwidth}{!}{
    \begin{tabular}{l c c c c c c c c | c}
    \toprule
        Loss  & Midjourney & SDv1.4 & SDv1.5 & ADM & GLIDE & Wukong & VQDM & BigGAN & mAcc\\
    \midrule
        Infonce/\added{$\tau$=0.5}   & 93.1 & 99.4 & 99.5 & 89.8 & 96.0 & 99.5 & 99.4 & 89.6 & 95.8\\
        Margin/cosine & 92.8 & 99.7 & 99.5 & 82.3 & 91.5 & 99.7 & 91.5 & 85.6 & 92.8\\
        Margin/euclidean & 94.8 & 98.5 & 98.3 & 94.7 & 96.1 & 98.6 & 98.5 & 98.0 & \textbf{97.2}\\
        \bottomrule \hline
    \end{tabular}
}
    \label{tab:choice}
\end{table*}

\begin{table*}[!t]
    \centering
    \caption{The impact of random patch replacement vs. fixed position replacement.}
    \resizebox{1.0\textwidth}{!}{
    \begin{tabular}{l c c c c c c c c | c}
    \toprule
        Position  & Midjourney & SDv1.4 & SDv1.5 & ADM & GLIDE & Wukong & VQDM & BigGAN & mAcc\\
    \midrule
        Upper Half & 87.1 & 99.7 & 99.6 & 83.9 & 94.5 & 99.7 & 99.2 & 96.0 & 95.0 \\
        Lower Half & 87.5 & 99.7 & 99.6 & 81.7 & 93.4 & 99.7 & 99.1 & 94.6 & 94.4 \\
        Left Half & 87.7 & 99.8 & 99.5 & 84.4 & 93.8 & 99.7 & 99.0 & 90.6 & 94.3 \\
        Right Half & 81.6 & 99.8 & 99.7 & 74.8 & 79.6 & 99.8 & 98.7 & 86.2 & 90.0 \\
        Random & 94.8 & 98.5 & 98.3 & 94.7 & 96.1 & 98.6 & 98.5 & 98.0 & \textbf{97.2}\\
        \bottomrule\hline
    \end{tabular}
}
  
    \label{tab:mix}
\end{table*}

\paragraph{Impact of patch size of randomized patch reconstruction.}
Our randomized patch reconstruction method reconstructs real images into fake counterparts where the patch size may influence performance. with patch size potentially influencing performance. Using patches of size $14 \times 14$ is intuitive, as it aligns with the token size used in Vision Transformers (ViTs). Tab.~\ref{tab:patchsize} illustrates that the training process is not sensitive to reconstruction patch size, and smaller patch sizes are preferred.

\begin{table*}[!t]
    \centering
    \caption{The impact of patch size of random patch replacement.}
    \resizebox{1.0\textwidth}{!}{
    \begin{tabular}{l c c c c c c c c | c}
    \toprule
        Patch Size  & Midjourney & SDv1.4 & SDv1.5 & ADM & GLIDE & Wukong & VQDM & BigGAN & mAcc\\
    \midrule
        112 & 99.6 & 99.3 & 92.3 & 92.7 & 99.5 & 95.8 & 99.5 & 94.3 & 96.6 \\
        56 & 99.4 & 99.2 & 93.8 & 90.8 & 99.4 & 95.8 & 99.1 & 97.2 & 96.8 \\
        28 & 98.7 & 98.3 & 95.7 & 95.4 & 98.8 & 97.5 & 98.5 & 98.1 & \textbf{97.6}\\
        14 & 94.8 & 98.5 & 98.3 & 94.7 & 96.1 & 98.6 & 98.5 & 98.0 & \underline{97.2}\\
        \bottomrule \hline
    \end{tabular}
}
    \label{tab:patchsize}
\end{table*}

\begin{table*}[!t]
    \centering
    \caption{The impact of random patch replacement vs. random patch dropout.}
\resizebox{1.0\textwidth}{!}{
    \begin{tabular}{l c c c c c c c c | c}
    \toprule
        Dropout Rate & Midjourney & SDv1.4 & SDv1.5 & ADM & GLIDE & Wukong & VQDM & BigGAN & mAcc\\
    \midrule
        0.10 & 71.3 & 99.9 & 99.8 & 67.9 & 82.7 & 99.8 & 97.3 & 77.0 & 88.5 \\
        0.15 & 94.3 & 98.7 & 98.6 & 87.9 & 90.4 & 98.7 & 98.5 & 90.9 & \underline{94.2} \\
        0.20 & 77.0 & 99.9 & 99.8 & 69.2 & 72.0 & 99.8 & 98.7 & 87.7 & 89.1 \\
        0.25 & 77.2 & 99.9 & 99.8 & 71.1 & 74.7 & 99.8 & 98.6 & 89.3 & 90.2 \\
        Replacement & 94.8 & 98.5 & 98.3 & 94.7 & 96.1 & 98.6 & 98.5 & 98.0 & \textbf{97.2}\\
\bottomrule \hline
    \end{tabular}
}
  
  \label{tab:dropout}
\end{table*}

\begin{figure}[t!]
    \centering
    \begin{adjustbox}{width=0.9\linewidth}
    \includegraphics{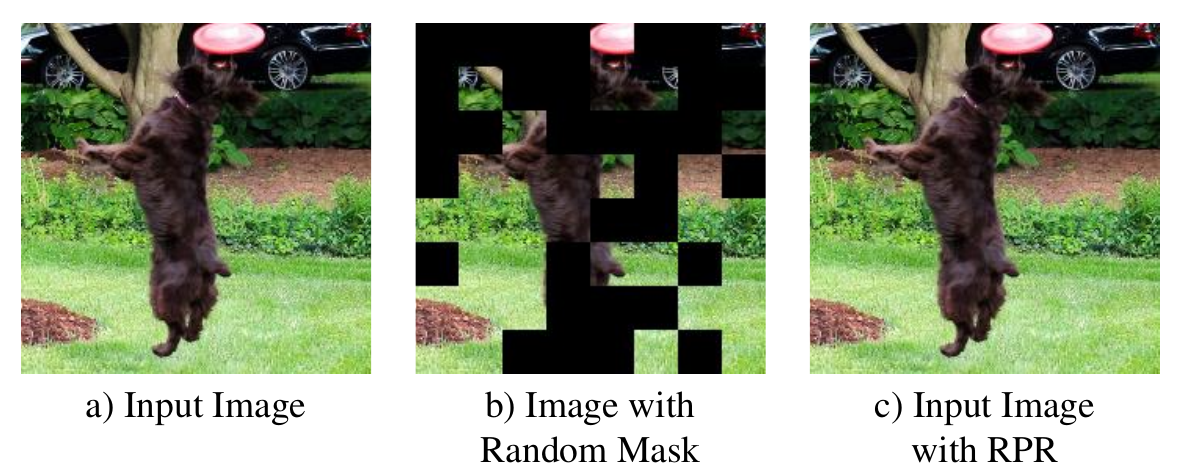}
    \end{adjustbox}
    \caption{Visual comparison between random patch dropout (masking) and reconstruction. It is evident that, by reconstruction, the overall visual appearance remains unchanged.}
    \label{fig:mask}
\end{figure}

\paragraph{Impact of randomized patch reconstruction vs. random patch dropout.}
Our randomized patch reconstruction method involves substituting fake image patches with their real counterparts, thereby compelling the model to exploit more artifacts from the remaining patches. An alternative approach shown in Fig.~\ref{fig:mask} is random patch dropout, in which certain patches are removed, resulting in images with fewer patches. Tab.~\ref{tab:dropout} illustrates the effectiveness of Randomized Patch Reconstruction, by comparing patch reconstruction with patch dropout at various dropout rates, with both methods employing patch-wise contrastive learning. The results indicate that, with an appropriate dropout ratio, patch dropout also achieves favorable performance, supporting our hypothesis that models tend to over-rely on a subset of patches. Dropout thus serves as a remedy for this issue. However, patch dropout underperforms patch reconstruction, possibly because patch reconstruction preserves the overall appearance and input domain of the image (i.e., a complete image rather than a masked one), thereby increasing the task's difficulty.

\section{More Visual Comparative Analysis}
\label{sec:comparative}
\begin{figure}[t!]
\centering
\subfloat[SDv1.4]{\includegraphics[height=2.7cm,width=3.4cm]{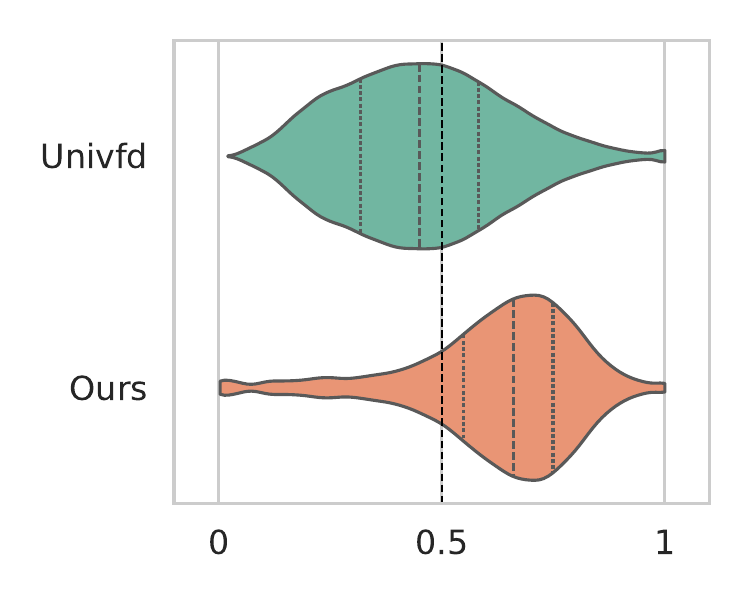}}
\subfloat[VQDM]{\includegraphics[height=2.7cm,width=3.4cm]{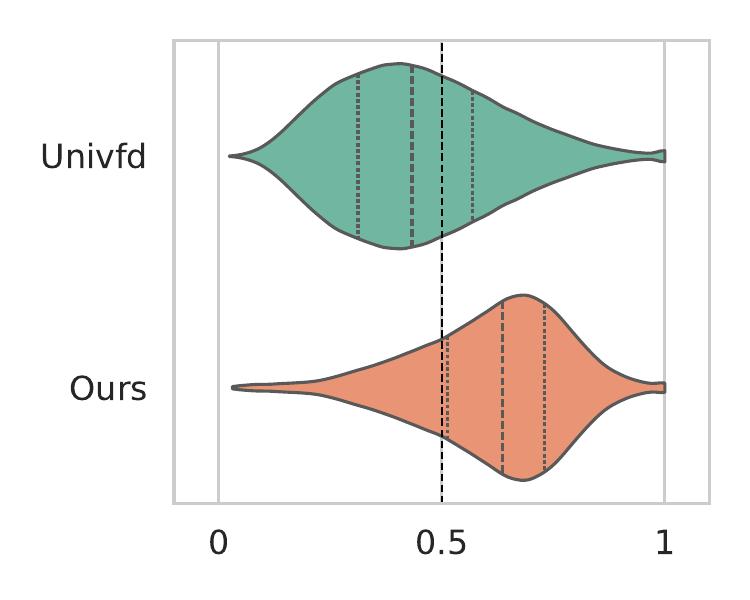}}
\subfloat[BigGAN]{\includegraphics[height=2.7cm,width=3.4cm]{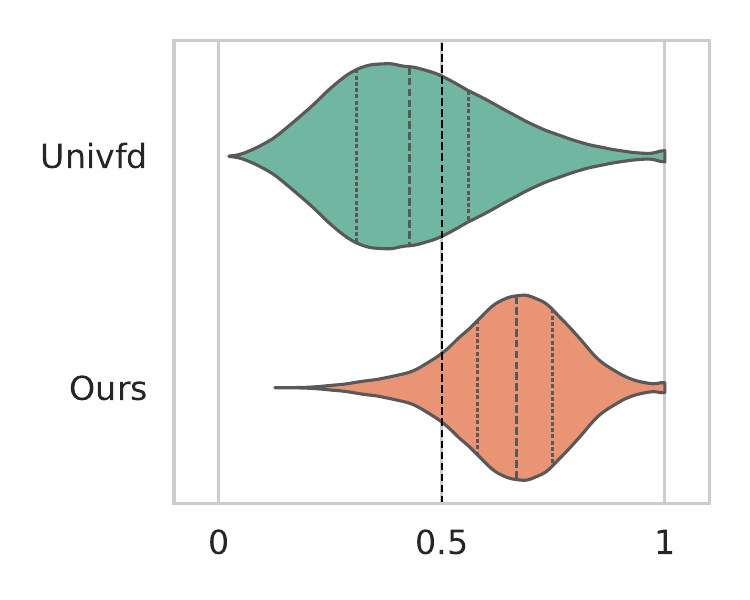}}
\subfloat[Glide]{\includegraphics[height=2.7cm,width=3.4cm]{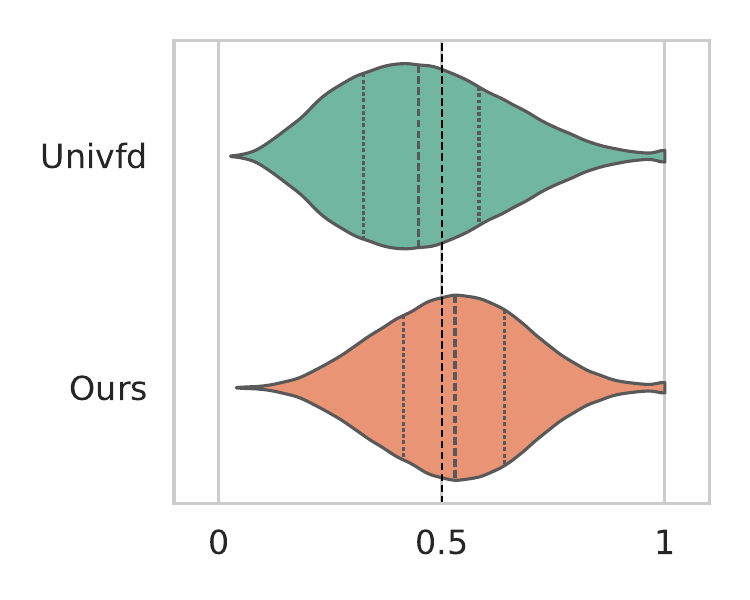}}
\caption{\added{CDE} distribution on different generators of ours and UnivFD.}
\label{fig:comparison}
\end{figure}

\paragraph{\added{CDE} distribution of different subsets of GenImage.}
To better analyze the models' ability to leverage all patches from an image, we use \added{CDE} to count the contribution of $\text{patch}_{(i,j)}$, which can be defined as the difference in logits at the $(i, j)$ position of an image before and after being masked.
Fig.~\ref{fig:comparison} illustrates the \added{CDE} distribution of UnivFD and our method. For better statistical analysis, we normalize the \added{CDE} values to a range $[0,1]$ using the exponential function $e^{TDE_{(i,j)} - TDE_{max}}$.
This normalization facilitates the measurement of differences between less dominant patches and the most dominant patches in the images. 
The figure demonstrates that a greater number of patches from our method are more uniform.

\begin{figure}[h]
    \centering
    \begin{adjustbox}{width=0.95\linewidth}
    \includegraphics{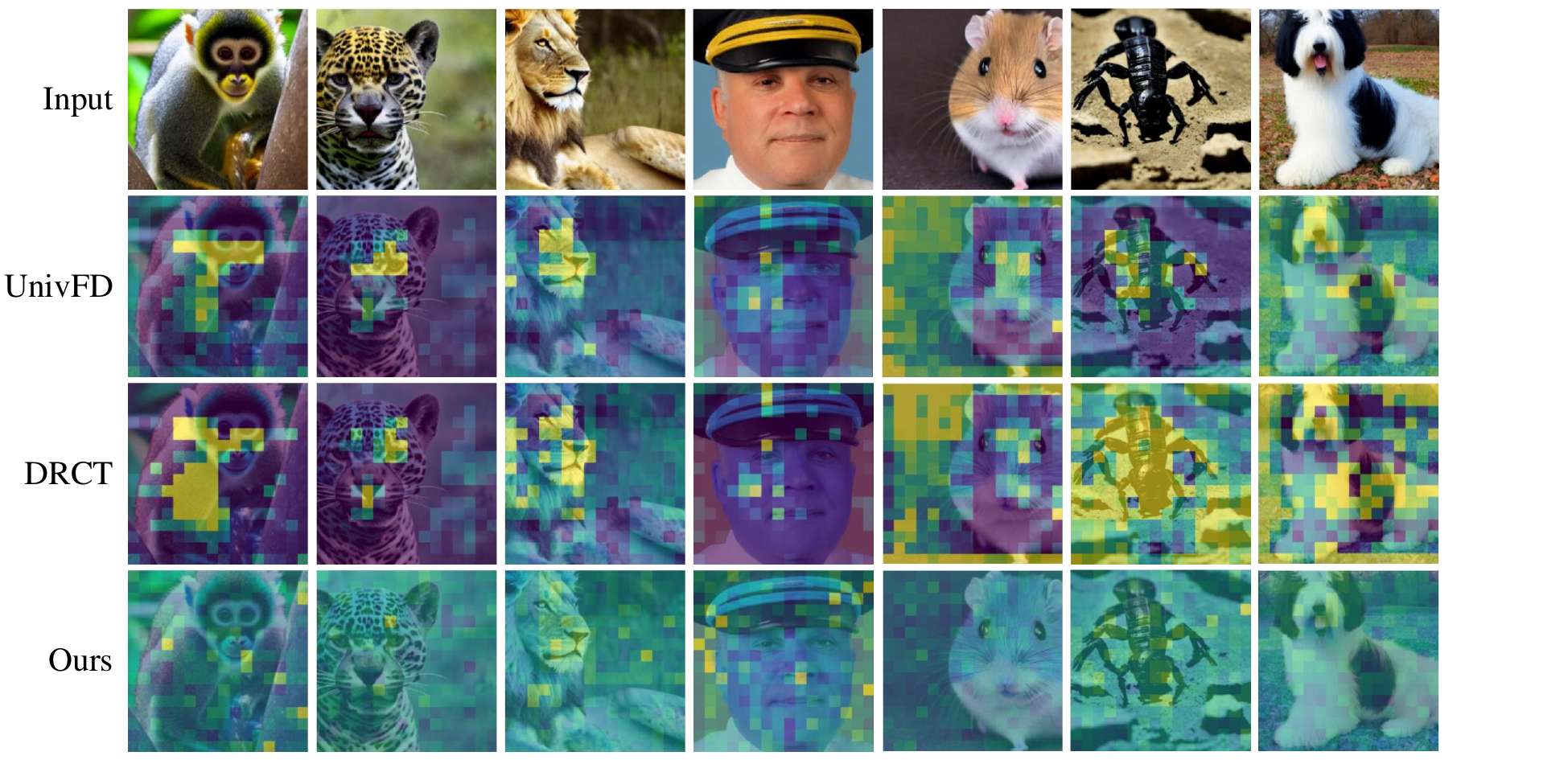}
    \end{adjustbox}
    \caption{Showcase of \added{CDE} map on SDv1.4. Images are sourced from GenImage~\citep{zhu2024genimage}.}
    \label{fig:tde_sd}
\end{figure}

\begin{figure}[h]
    \centering
    \begin{adjustbox}{width=0.95\linewidth}
    \includegraphics{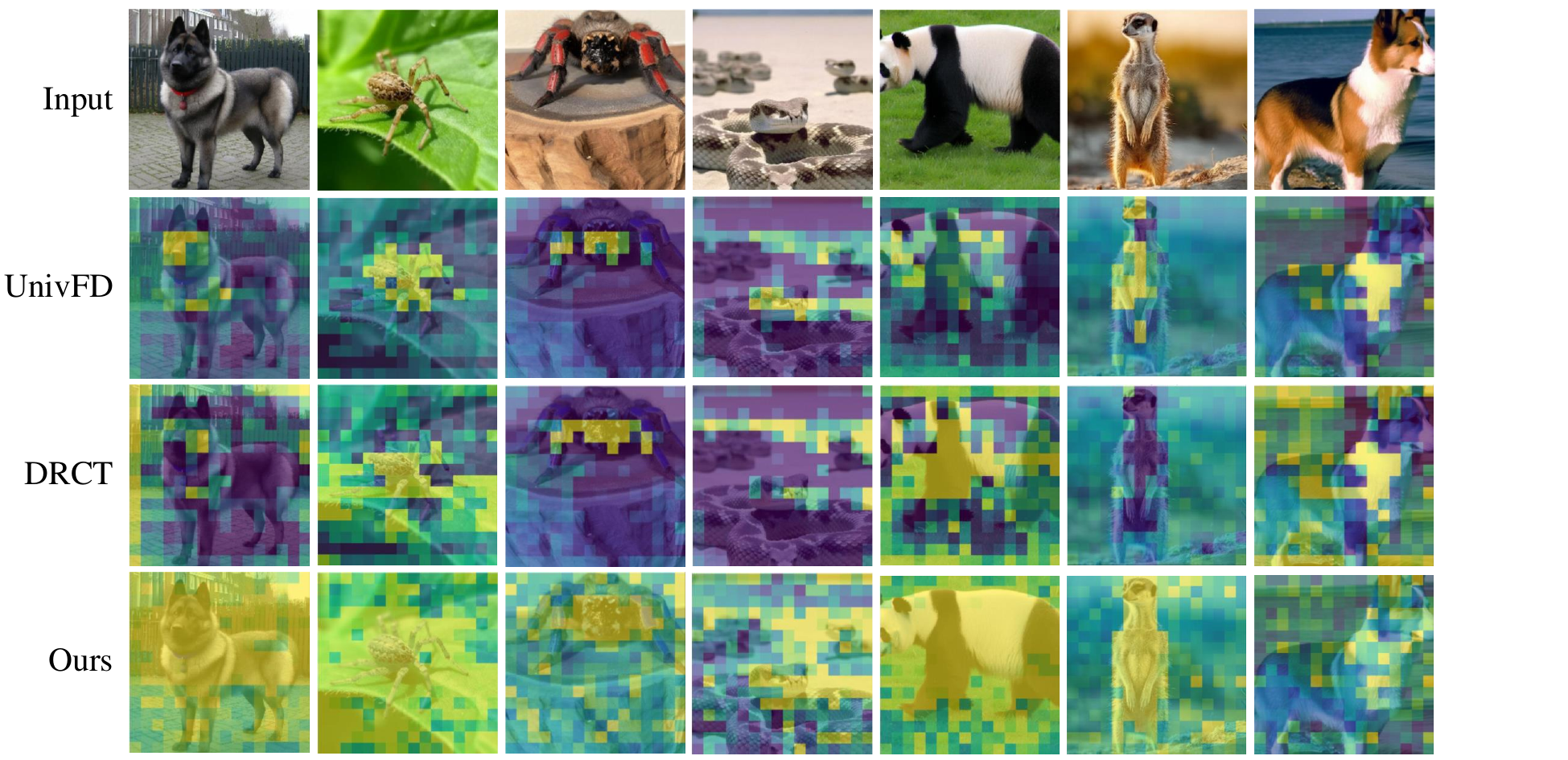}
    \end{adjustbox}
    \caption{Showcase of \added{CDE} map on Midjourney. Images are sourced from GenImage~\citep{zhu2024genimage}.}
    \label{fig:tde_mj}
\end{figure}

\begin{figure}[h]
    \centering
    \begin{adjustbox}{width=0.95\linewidth}
    \includegraphics{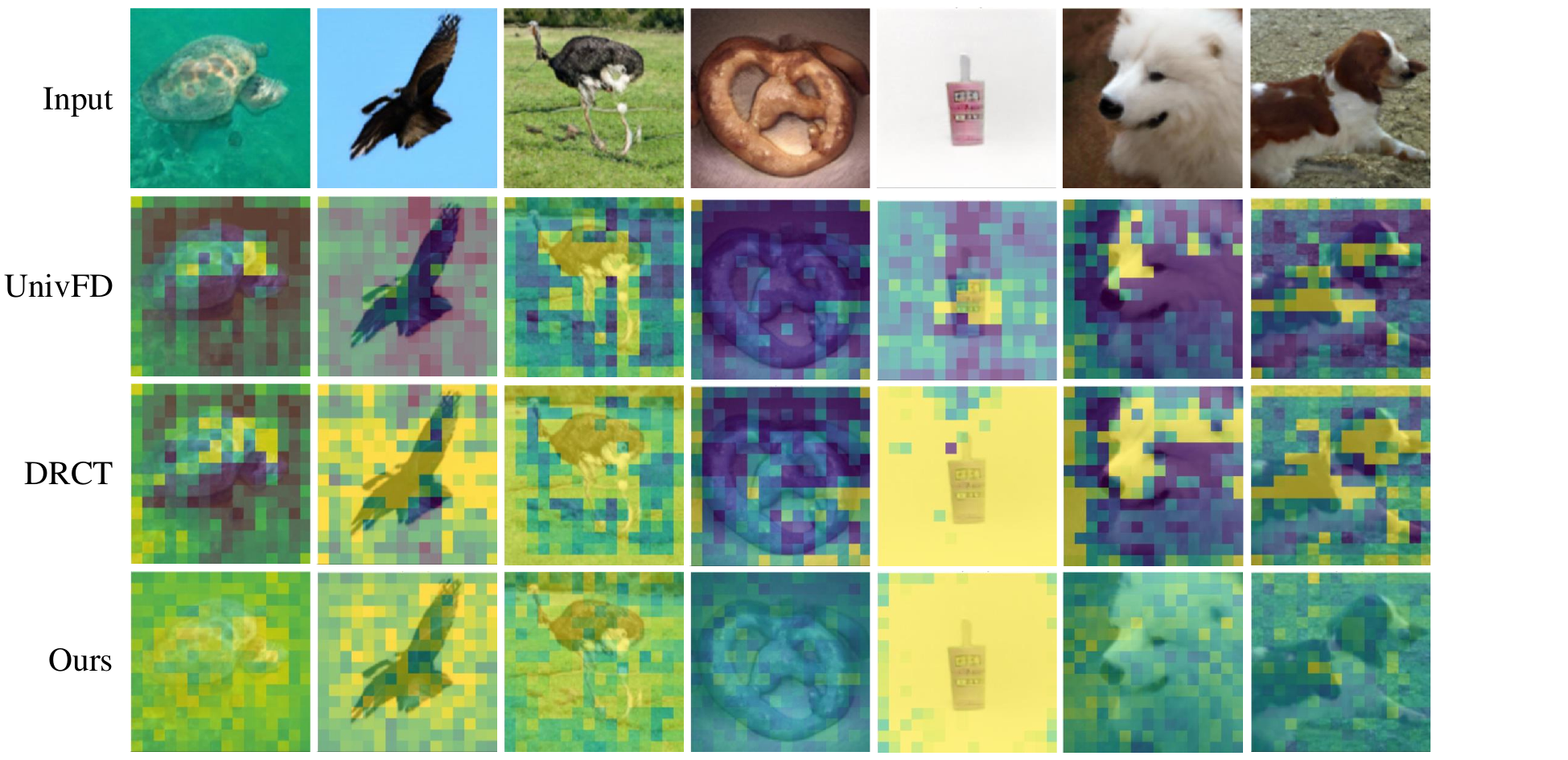}
    \end{adjustbox}
    \caption{Showcase of \added{CDE} map on BigGAN. Images are sourced from GenImage~\citep{zhu2024genimage}.}
    \label{fig:tde_gan}
\end{figure}

\begin{figure}[h]
    \centering
    \begin{adjustbox}{width=0.95\linewidth}
    \includegraphics{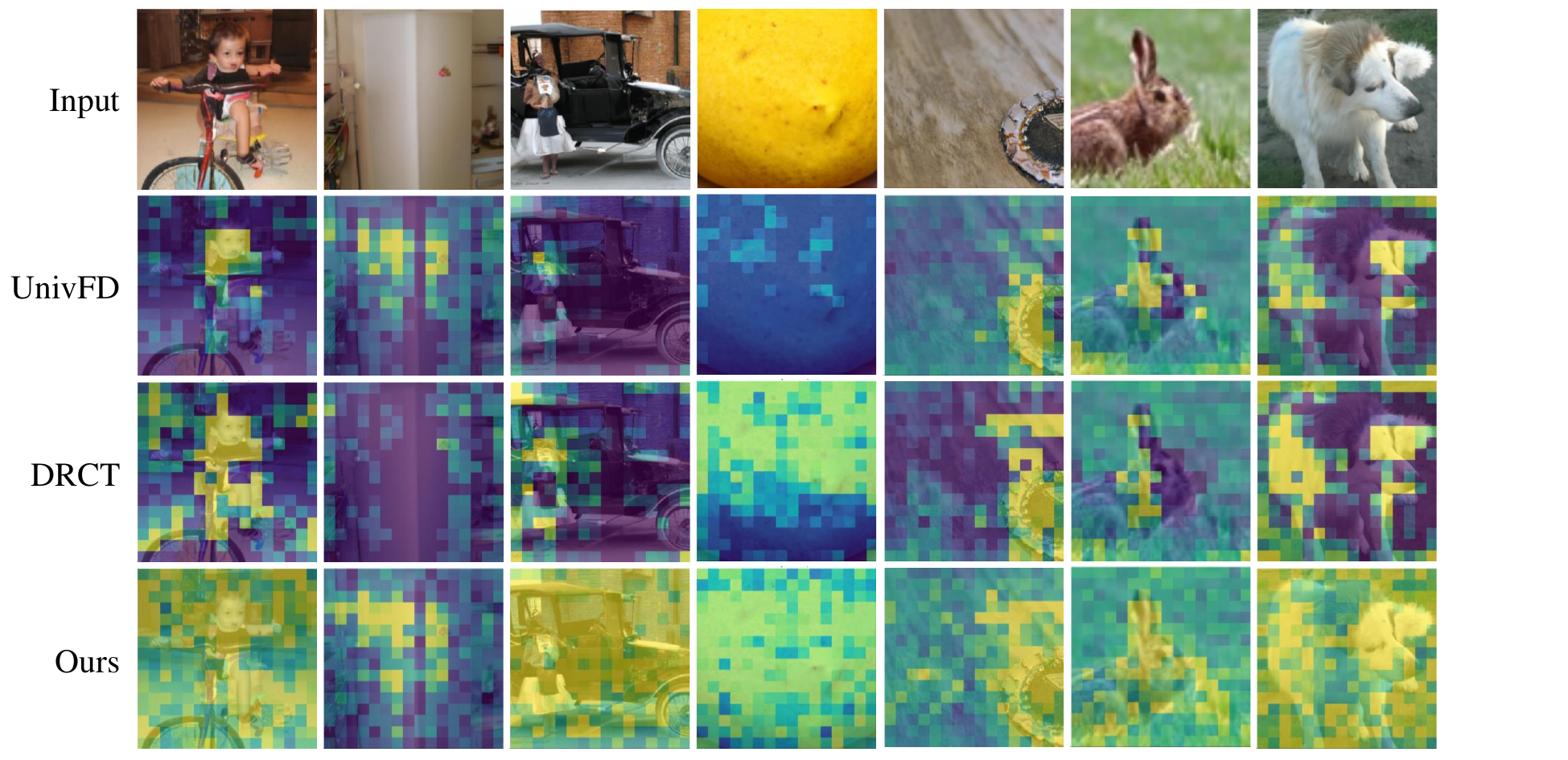}
    \end{adjustbox}
    \caption{Showcase of \added{CDE} map on ADM. Images are sourced from GenImage~\citep{zhu2024genimage}.}
    \label{fig:tde_adm}
\end{figure}

\begin{figure}[h]
    \centering
    \begin{adjustbox}{width=0.95\linewidth}
    \includegraphics{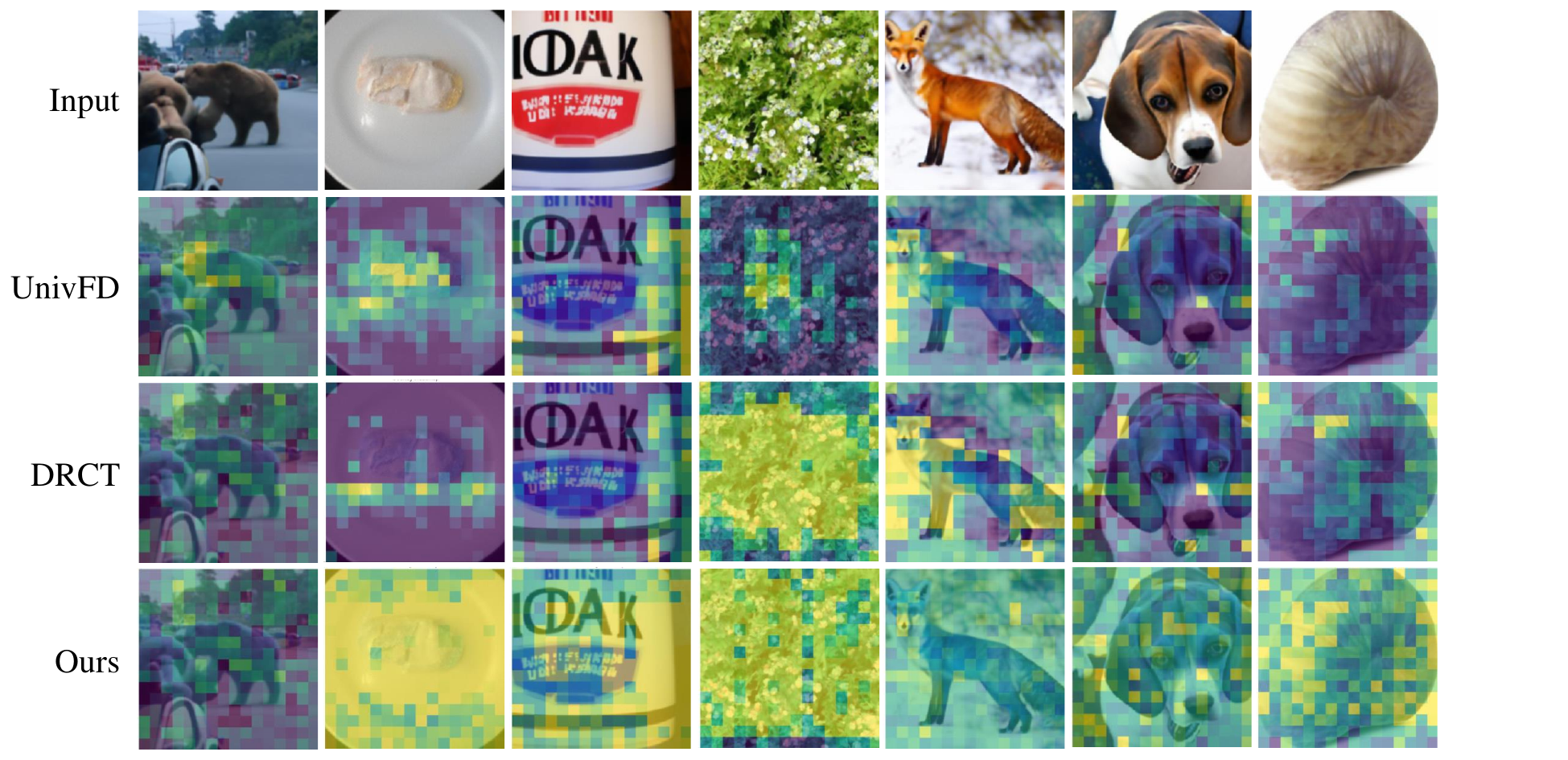}
    \end{adjustbox}
    \caption{Showcase of \added{CDE} map on Glide. Images are sourced from GenImage~\citep{zhu2024genimage}.}
    \label{fig:tde_glide}
\end{figure}

\begin{figure}[h]
    \centering
    \begin{adjustbox}{width=0.95\linewidth}
    \includegraphics{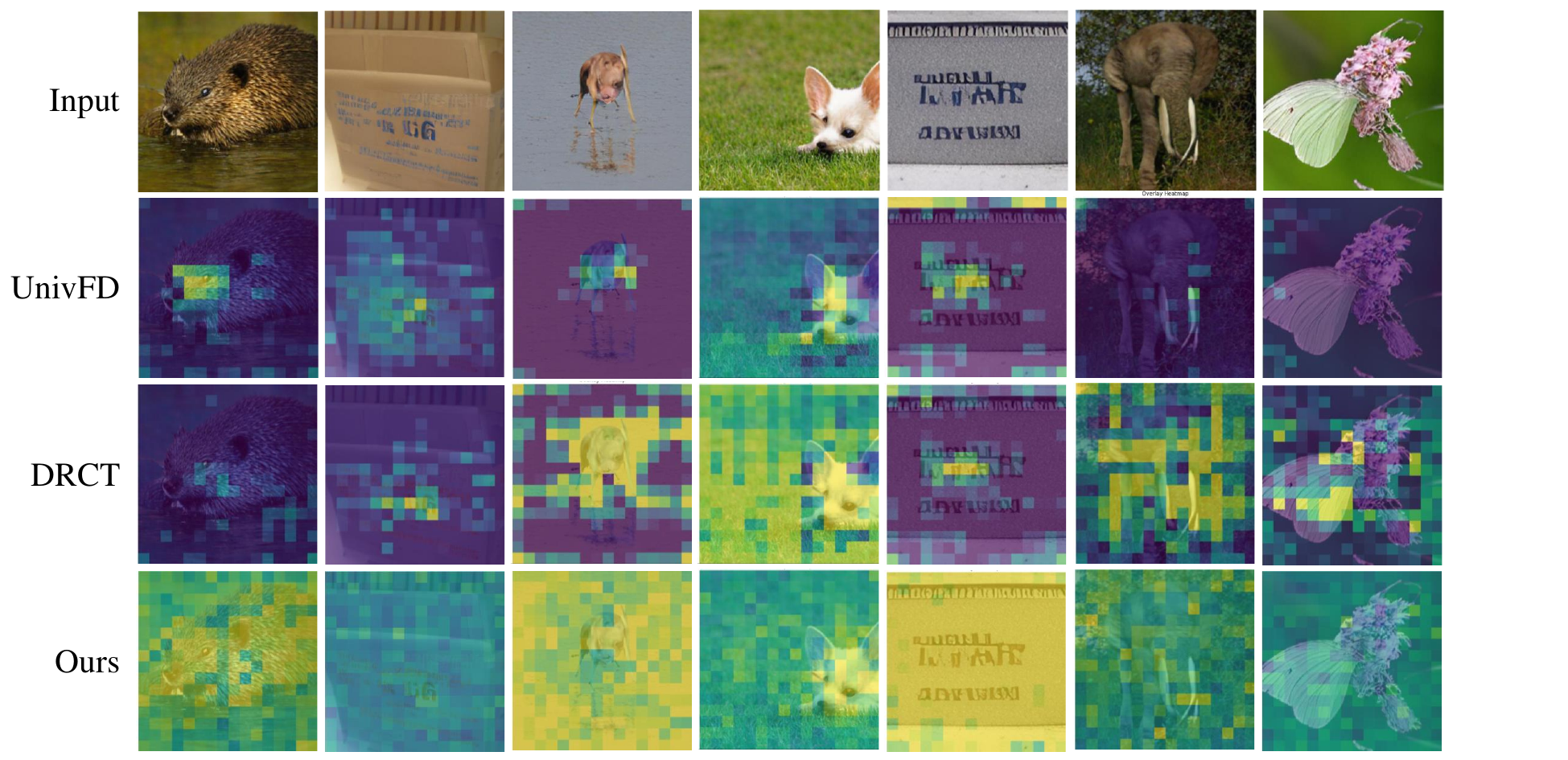}
    \end{adjustbox}
    \caption{Showcase of \added{CDE} map on VQDM. Images are sourced from GenImage~\citep{zhu2024genimage}.}
    \label{fig:tde_vqdm}
\end{figure}

\paragraph{Visual showcase of \added{CDE} distribution of different subsets on GenImage.}
To better showcase our model's better ability to leverage all patches from an image, we present a visual analysis of \added{CDE} across various subsets of the GenImage dataset. 
The GenImage dataset is divided into multiple subsets, each representing distinct image generation methods. 
These subsets include GAN-based models such as BigGAN, and diffusion-based models, including Stable Diffusion, VQDM, and ADM. 
Due to space limitations in the main text, we showcased limited images; here, we present most subset models of GenImage: the diffusion-based Stable Diffusion v1.4 (Fig.~\ref{fig:tde_sd}), the closed-source Midjourney (Fig.~\ref{fig:tde_mj}), and the GAN-based BigGAN (Fig.~\ref{fig:tde_gan}). The rest diffusion-based model are from Fig.~\ref{fig:tde_adm} to Fig.~\ref{fig:tde_vqdm}.We use CLIP as backbone for our visualization.

\begin{figure}[h]
    \centering
    \begin{adjustbox}{width=0.95\linewidth}
    \includegraphics{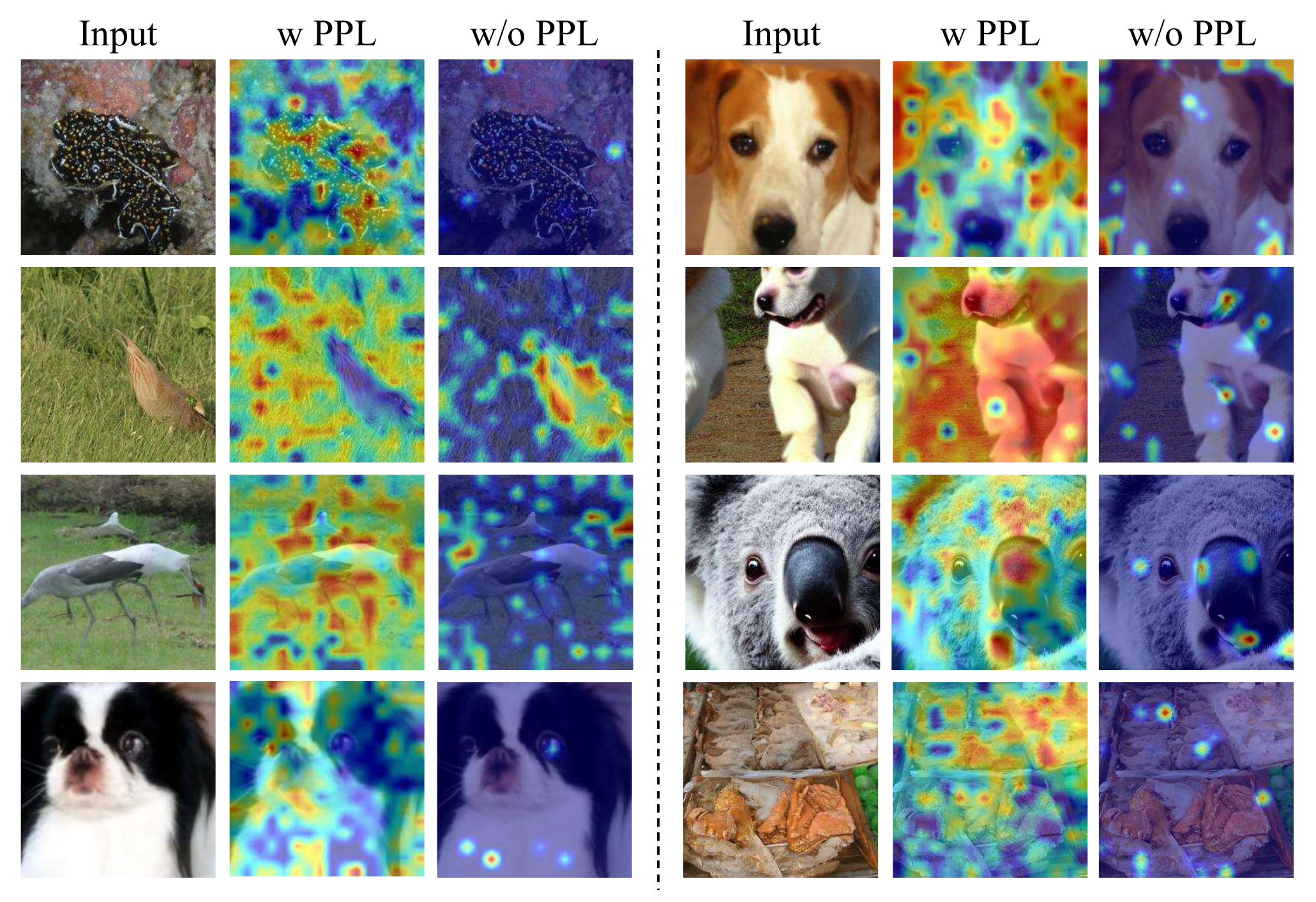}
    \end{adjustbox}
    \caption{Showcase of GradCAM map of PPL vs naive LoRA-tuning on GenImage. Images are sourced from GenImage~\citep{zhu2024genimage}.}
    \label{fig:111}
\end{figure}

\paragraph{Visual showcase of GradCAM  on GenImage.}
\added{To provide a more intuitive understanding of how our method alters the model's decision-making process, we employ Grad-CAM to visualize the attention maps of the baseline (naive LoRA-tuning) versus our proposed PPL using a CLIP backbone. As illustrated in Fig.~\ref{fig:111}, the baseline model tends to overfit to sparse, highly localized regions while ignoring the rest of the image. This confirms the prevalence of "Few-Patch Bias" in standard training. In contrast, our method demonstrates a significantly broader spatial coverage, attending to a diverse range of forgery traces across the entire image.}

\begin{figure}[h]
    \centering
    \begin{adjustbox}{width=0.95\linewidth}
    \includegraphics{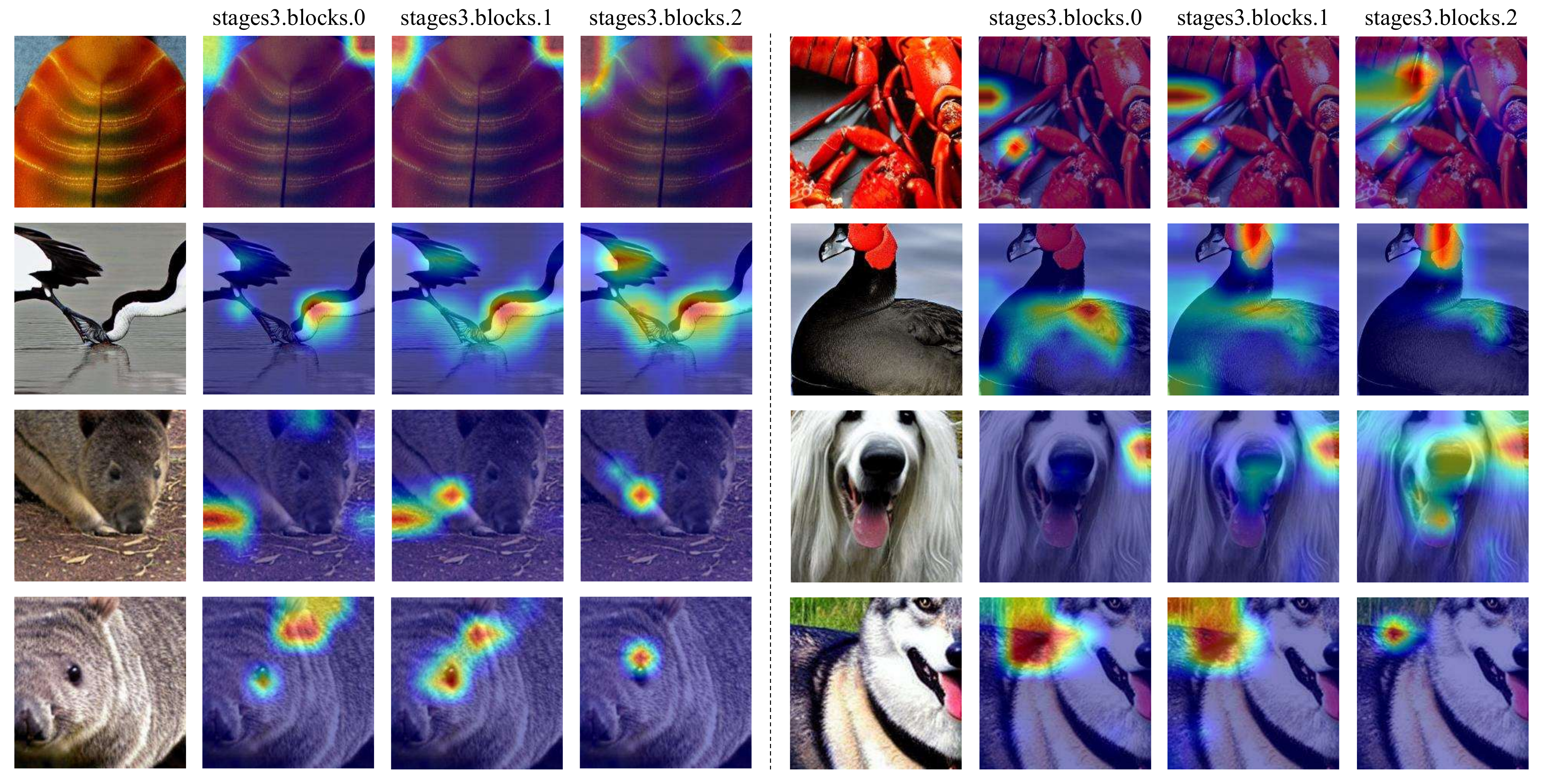}
    \end{adjustbox}
    \caption{Showcase of GradCAM map on last 3 blocks of ConvNeXt. Images are a sourced from GenImage~\citep{zhu2024genimage}.}
    \label{fig:222}
\end{figure}

\added{To further investigate the universality of these biases, we extend our visualization to CNN-based architectures. We analyze the activation maps of a ConvNeXt-Base detector with a specific focus on the evolution within the deepest blocks (Stage 3, Block 0,1,2) which are most closely related to decision-making according to \citep{zeiler2014visualizing}. Fig.~\ref{fig:222} shows that`Few-Patch Bias' is not exclusive to ViTs but is a fundamental characteristic of the decision layers in modern CNNs, further justifying the necessity of our proposed learning strategy.}

\section{The Use of Large Language Models (LLMs)}
In this paper, we only use the large language model to help polish our text. The large language model has no role in the research conception.

\end{document}